\documentclass[11pt]{article}

\usepackage[final]{acl}

\usepackage{times}
\usepackage{latexsym}

\usepackage[T1]{fontenc}

\usepackage[utf8]{inputenc}

\usepackage{microtype}

\usepackage{inconsolata}

\usepackage{graphicx}

\newcommand{\resourcelinks}{%
  \par\smallskip
  {\small
  \setlength{\tabcolsep}{2pt}%
  \begin{tabular}{@{}rl@{}}
    \raisebox{-0.15em}{\includegraphics[height=1.1em]{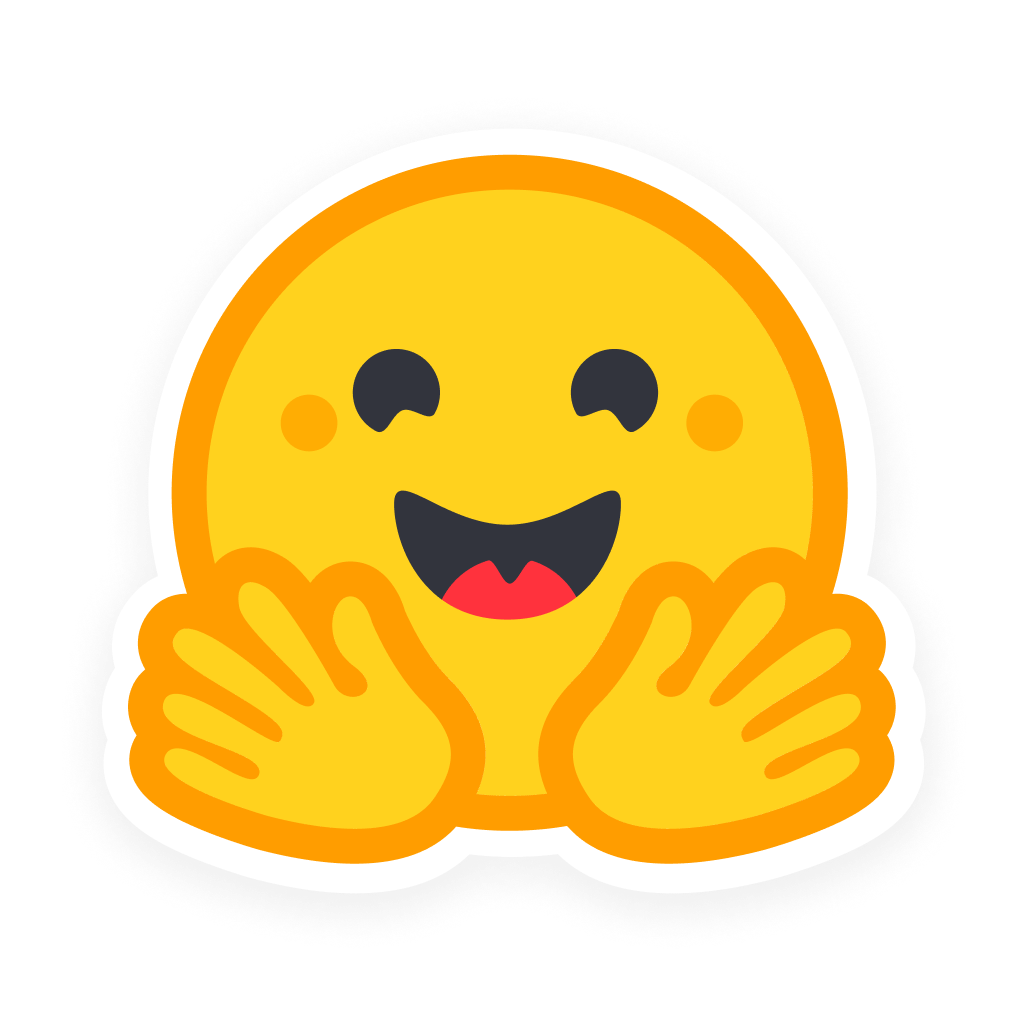}} &
    \textbf{Dataset: }
    \href{https://huggingface.co/datasets/llm-jp/HakushoBench}{llm-jp/HakushoBench}
    \\[0.3em]
    \raisebox{-0.15em}{\includegraphics[height=1.1em]{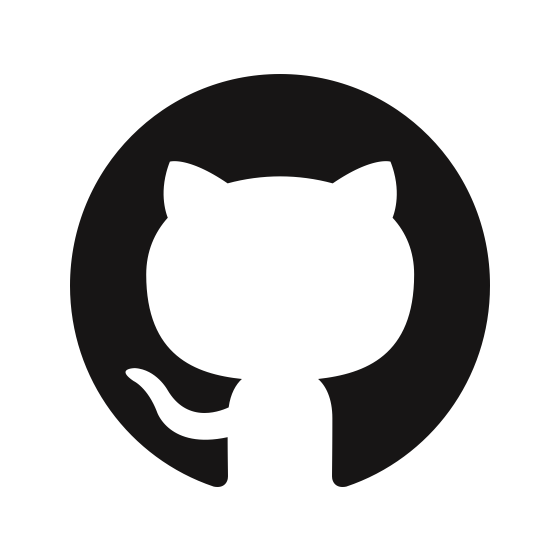}} &
    \textbf{Code: }
    \href{https://github.com/llm-jp/simple-evals-mm}{github.com/llm-jp/simple-evals-mm}
  \end{tabular}
  }
  \par
}

\usepackage{xurl}            %
\usepackage{hyperref}       %
\usepackage{booktabs}       %
\usepackage{amsfonts}       %
\usepackage{nicefrac}       %
\usepackage[table]{xcolor}      %

\usepackage{lineno}

\definecolor{darkblue}{rgb}{0, 0, 0.5}

\usepackage[whole]{bxcjkjatype}

\usepackage{wrapfig}

\hypersetup{
	colorlinks=true,
    citecolor=blue,
    linkcolor=blue,
    urlcolor=blue,
	pdfborder={0 0 0},
}

\usepackage{pifont}
\newcommand{\cmark}{\ding{51}}
\newcommand{\xmark}{\ding{55}}

\usepackage{mathtools}

\usepackage{multirow}

\usepackage{listings}
\usepackage{tcolorbox}
\tcbuselibrary{skins, breakable, listings}
\usepackage{tikz}
\usepackage{comment} %

\newtcolorbox[auto counter, number within=section]{Prompt}[2][]{%
  colback=white, %
  width=\linewidth, %
  arc=3mm,
  boxrule=0.8mm, %
  title=\large #2, %
  breakable, %
  fonttitle=\small, %
  fontupper=\footnotesize, %
  #1 %
}

\newtcolorbox{SlimCode}{
  enhanced,
  colback=gray!20,
  colframe=gray!50,
  boxrule=0.4pt,
  arc=2mm,
  outer arc=2mm,
  left=2mm,
  right=2mm,
  top=1mm,
  bottom=1mm,
  boxsep=2pt,
  breakable,
  listing options={
    basicstyle=\ttfamily\scriptsize,  %
    breaklines=true,
    columns=fullflexible,
  },
  before skip=5pt,   %
  after skip=5pt     %
}

\AtBeginEnvironment{SlimCode}{\footnotesize}

\newcommand{\method}{HakushoBench}

\usepackage{enumitem}

\title{HakushoBench: A Japanese Chart and Table VQA Benchmark from Governmental White Papers}

\newcommand{\aspace}{\hspace{0.7em}}

\author{
    Issa Sugiura$^{\heartsuit,\spadesuit}$  \aspace
    Shuhei Kurita$^{\diamondsuit,\spadesuit}$  \aspace
    Yusuke Oda$^{\spadesuit}$ \aspace
    Naoaki Okazaki$^{\heartsuit,\spadesuit,\clubsuit}$
    \\
    $\heartsuit$Institute of Science Tokyo \aspace
    $\spadesuit$NII LLMC \aspace
    $\diamondsuit$NII \aspace
    $\clubsuit$AIST
}

\begin{document}
\maketitle
\begin{abstract}
Understanding chart and table images is essential for applying vision-language models (VLMs) to real-world document understanding.
While English benchmarks have advanced rapidly, non-English counterparts remain scarce, leaving it unclear whether this progress generalizes across languages. A key obstacle is the difficulty of collecting realistic and diverse non-English chart and table images at scale. To address this, we leverage governmental white papers as a source for benchmark construction, as they contain naturally occurring charts and tables across diverse formats and domains and are freely accessible in many countries.
As a first instantiation, we introduce HakushoBench, a Japanese chart and table VQA benchmark built from 33 governmental white papers. HakushoBench contains 2,053 images spanning over 10 image types, with manually annotated and independently verified QA pairs designed to assess holistic understanding of charts and tables rather than local visual cues alone.
Experiments across a broad range of VLMs show that HakushoBench is substantially harder than the existing Japanese benchmark and remains challenging for open-weight models: sub-10B open-weight models reach at most 58.6\% accuracy, and even the flagship open-weight model Qwen3.5-397B-A17B trails Gemini~3~Pro by 8.1 points (85.8\% vs.\ 93.9\%), highlighting substantial room for improvement in complex chart and table understanding. We release our dataset and code.
\resourcelinks
\end{abstract}

\begin{figure}[t]
  \centering
  \includegraphics[width=\linewidth]{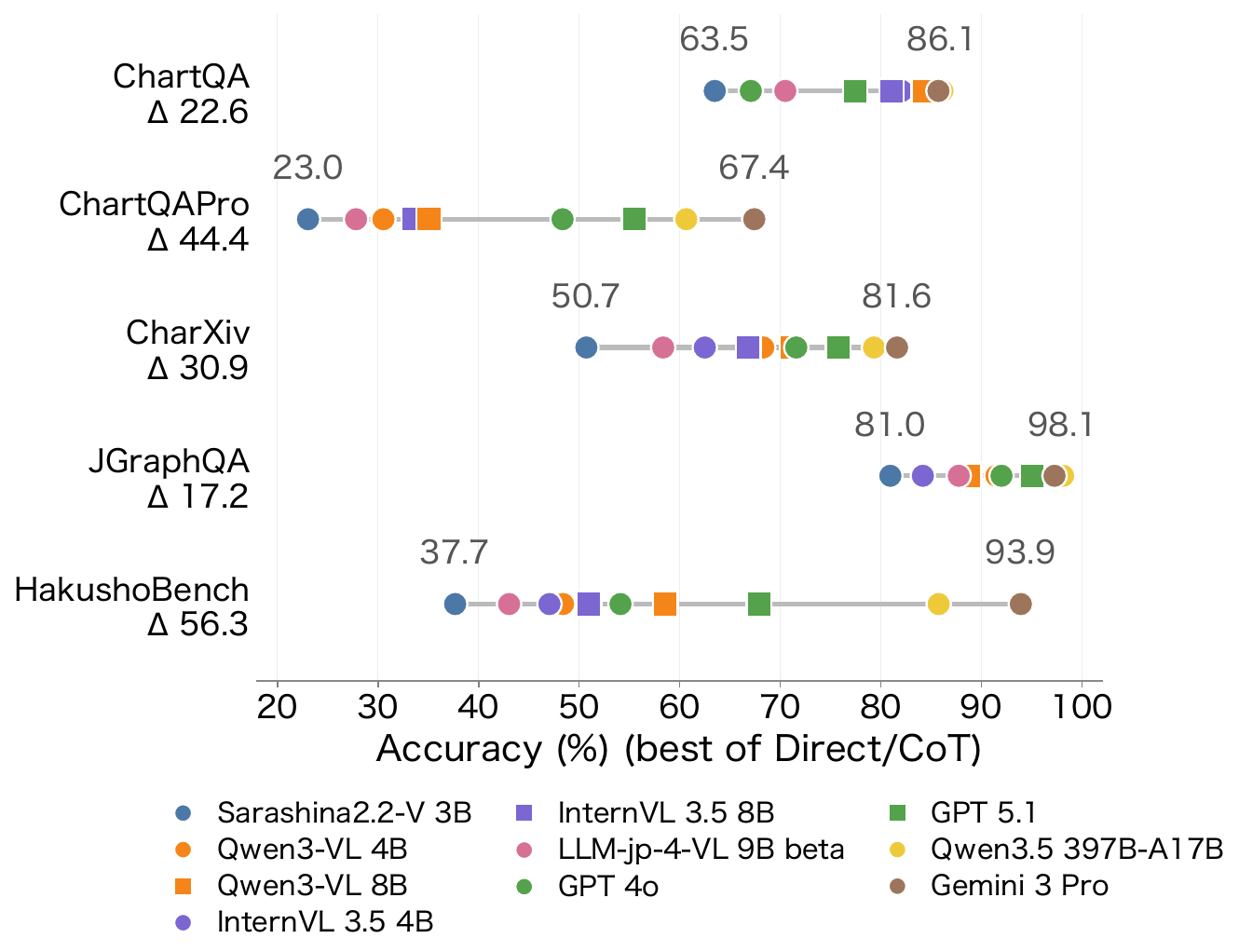}
  \caption{Score distribution of the evaluated models on each benchmark (best of Direct/CoT; $\Delta$ = max $-$ min). Every model scores lower on \method{} than on the existing Japanese benchmark JGraphQA, where scores cluster above 80\%.}
  \label{fig:bench_range}
\end{figure}

\begin{figure*}[t]
  \includegraphics[width=\linewidth]{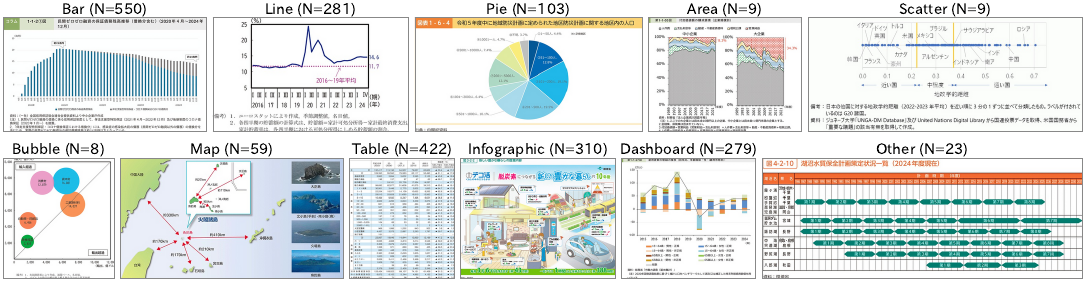}
  \caption{Diversity of image types in \method{}. One randomly sampled example is shown for each image type included in \method{}. \method{} contains diverse, well-designed chart and table images that present rich information in an accessible and visually understandable manner for general readers of government white papers.}
  \label{fig:showcase_by_type}
\end{figure*}

\section{Introduction}

Vision-language models (VLMs)~\citep{liu2023llava,openai2024gpt4ocard,bai2025qwen3vl,kimiteam2025kimivl} have rapidly advanced as general-purpose models capable of solving a wide range of vision-language tasks~\citep{Antol2015VQA,masry2022chartqa,lu2024mathvista,xie2024osworld}. 
Among these tasks, chart and table visual question answering (VQA) represents a fundamental capability for VLMs, as charts and tables are ubiquitous across a wide range of real-world documents, including public reports, financial documents, and scientific articles, often conveying information that cannot be expressed in text alone~\citep{masry2022chartqa,kantharaj2022chart-to-text}.

To evaluate chart and table understanding in VLMs, numerous benchmarks have been developed. Early datasets such as ChartQA~\citep{masry2022chartqa} and PlotQA~\citep{methani2020plotqa} mainly focused on relatively simple chart types and questions requiring straightforward extraction of numerical values or labels. As VLM performance rapidly improved, these benchmarks became increasingly saturated and less effective at distinguishing model capabilities~\citep{ho2025rosettastone}. 
More recent benchmarks, such as ChartQAPro~\citep{masry2025chartqapro} and CharXiv~\citep{wang2024charxiv}, address this limitation by introducing more diverse real-world chart images and more challenging question-answer pairs.

However, existing chart and table benchmarks are heavily biased toward English-centric visual conventions and document styles~\citep{masry2022chartqa,wang2024charxiv,masry2025chartqapro}. 
Real-world chart understanding varies across languages and cultures, affecting visual composition, textual structure, and reasoning requirements~\citep{tang-etal-2025-mtvqa,xu2026polychartqa}. 
For example, chart and table images may differ in geographic conventions used in map-based figures, language-specific terminology in tables, writing direction (e.g., mixed vertical and horizontal text), and overall information density and layout structure~\citep{wei2025deepseekocr,sasagawa2025vertical,onami2024jdocqa}. As a result, strong performance on existing English benchmarks does not imply robust multilingual chart understanding~\citep{globerson2024nofilter}.

In Japanese, JGraphQA~\citep{jgraphqa} is currently the primary benchmark for chart and table understanding. However, it consists of only around 200 instances with limited visual diversity and relatively simple questions, leading to performance saturation where even 3B-scale VLMs already achieve over 80\% accuracy (Figure~\ref{fig:bench_range}).

To address these limitations, we leverage governmental white papers (\textit{Hakusho}, 白書, in Japanese) as an underexplored yet valuable source for chart and table benchmark construction. These reports contain large amounts of real-world charts and tables spanning diverse domains and visual formats, and are publicly available across many countries and languages~\citep{egovjp,usgov2026economic}.

As a first instantiation of this approach, we introduce \textbf{\method{}}, a challenging benchmark for Japanese chart and table understanding constructed from 33 white papers published by Japanese governmental agencies. The benchmark contains 2,053 unique images spanning more than 10 image types, with manually annotated QA pairs designed to be challenging, including questions that require integrating information across the entire image and multi-hop reasoning beyond simple data extraction.

We evaluate a broad suite of open-weight and proprietary VLMs on \method{} and compare against existing Japanese and English chart and table benchmarks. \method{} is substantially harder than JGraphQA: sub-10B open-weight models reach at most 58.6\% accuracy (Qwen3-VL~8B~\citep{bai2025qwen3vl}), and even the flagship open-weight model Qwen3.5-397B-A17B~\citep{qwen35blog}, which matches or exceeds Gemini~3~Pro on JGraphQA and ChartQA, trails it by 8.1 points on \method{} (85.8\% vs.\ 93.9\%). The resulting 56-point spread across models distinguishes them far more clearly than JGraphQA, where scores cluster between 81\% and 98\%.

\begin{table}[t]
\centering
\small
\setlength{\tabcolsep}{3pt}
\begin{tabular}{lccc}
\toprule
\textbf{Benchmark} & \textbf{Real} & \textbf{Image Types} & \textbf{Lang.} \\
\midrule
PlotQA{\tiny~\citep{methani2020plotqa}}        & \xmark & 3   & EN \\
ChartQA{\tiny~\citep{masry2022chartqa}}       & \cmark & 3  & EN \\
ChartQAPro{\tiny~\citep{masry2025chartqapro}}    & \cmark  & >10  & EN \\
CharXiv {\tiny~\citep{wang2024charxiv}}      & \cmark & >10  & EN \\
\midrule
JGraphQA{\tiny~\citep{jgraphqa}}      & \cmark  & 5   & JA \\
\method{} (Ours)     &\cmark & >10 & JA \\
\bottomrule
\end{tabular}
\caption{Comparison of chart and table VQA benchmarks. ``Real'' indicates that images are collected from real-world sources rather than programmatically generated. \method{} provides realistic and visually diverse Japanese chart and table VQA.}
\label{tab:comparison_benchmark}
\end{table}

\begin{figure*}[t]
  \includegraphics[width=\linewidth]{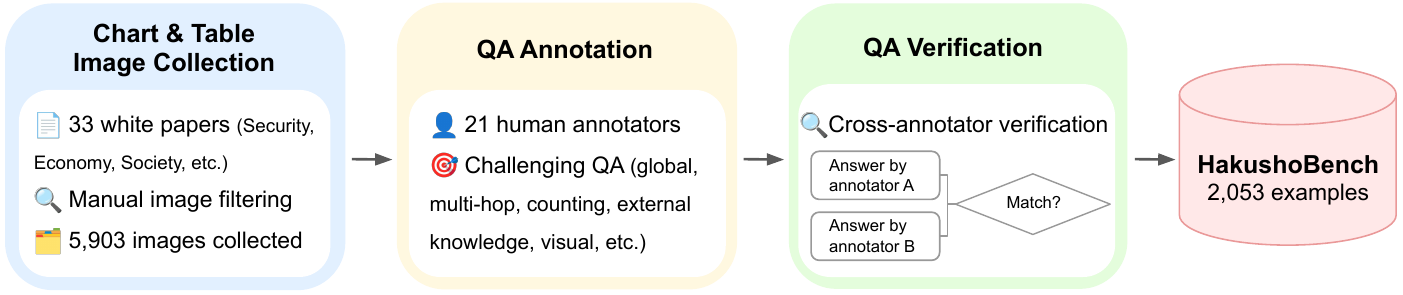}
  \caption{Construction pipeline of \method{}. Chart and table images are collected from 33 Japanese white papers and filtered to 5,903 candidates. Annotators then create one high-difficulty QA pair per image, followed by independent verification, yielding 2,053 VQA pairs.}
  \label{fig:hakushobench_pipeline}
\end{figure*}

\section{Related Work}
\paragraph{Chart and table VQA benchmarks.}
As vision-language models have become increasingly capable and general-purpose~\citep{liu2023llava,bai2025qwen3vl}, chart and table VQA benchmarks have also evolved toward greater diversity and more challenging reasoning tasks.

Early datasets such as FigureQA~\citep{ebrahimi2018figureqa} and DVQA~\citep{Kafle2018DVQA} focused on synthetic bar charts, while later work such as PlotQA~\citep{methani2020plotqa} expanded chart types and task complexity. To bridge the gap with real-world documents, ChartQA~\citep{masry2022chartqa} introduced human-annotated QA pairs on real chart images collected from online platforms such as Statista~\citep{statista2026statista} and Our World in Data~\citep{owid2026our_world_in_data}, which present diverse topics across a broad range of domains. 

More recent benchmarks have further expanded the scope of image sources: ChartQAPro~\citep{masry2025chartqapro} collected charts from a larger set of online platforms, covering more varied types such as dashboards and infographics, while CharXiv~\citep{wang2024charxiv} extracted chart images from papers across eight arXiv categories (including Computer Science, Economics, and Physics) to construct a natural and challenging benchmark.
ChartMuseum~\citep{tang2026chartmuseum} further targets questions that require visual rather than textual reasoning, and MTabVQA~\citep{singh-etal-2025-mtabvqa} evaluates multi-hop reasoning over multiple rendered tables.
Despite these advances, chart and table benchmark development has primarily focused on English.

\paragraph{Japanese chart and table VQA benchmarks.}
In Japanese, several benchmarks have been proposed for document understanding. JDocQA~\citep{onami2024jdocqa} built a document VQA benchmark from diverse PDFs published by Japanese public institutions such as municipal offices, and BusinessSlideVQA~\citep{stockmark2025businessslidevqa} constructed a slide understanding benchmark from business slides released by Japanese companies. 

The most closely related prior work, JGraphQA~\citep{jgraphqa}, focuses on Japanese chart and table question answering and is constructed from Investor Relations (IR) presentation slides of Japanese companies. However, this source domain is relatively narrow, and the benchmark contains only around 200 examples. Furthermore, following the QA design of ChartQA, questions are comparatively simple and focused primarily on data extraction and basic arithmetic rather than complex reasoning. As a result, even 3B-scale VLMs already achieve over 80\% accuracy, suggesting that the benchmark is becoming saturated.

Table~\ref{tab:comparison_benchmark} summarizes existing chart and table VQA benchmarks alongside \method{}. In contrast to prior work, \method{} leverages governmental white papers as an image source, enabling a realistic, visually diverse, and challenging Japanese chart and table VQA benchmark.

\section{Construction of \method{}}

\method{} is constructed through a three-stage pipeline, illustrated in Figure~\ref{fig:hakushobench_pipeline}. 

\subsection{Chart and Table Image Collection from Japanese White Papers}
\label{subsec:image_collection}

\paragraph{White papers as a data source.}
White papers published by Japanese governmental agencies summarize official statistics and policy analyses on national topics such as defense, energy, welfare, and education for a general readership. As a result, the images they contain are professionally designed, information-dense, and visually diverse across a wide range of domains, making them a valuable source for evaluating the generalizability of VLMs across heterogeneous chart and table types.
Furthermore, since many governments around the world similarly publish official white papers~\citep{usgov2026economic}, the same source-level methodology could be applied to other languages. We demonstrate it for Japanese in this paper and leave multilingual instantiation to future work.

\paragraph{Scope and edition selection.}
We collect chart and table images from such white papers, which are publicly available through the Japanese government's e-Gov portal~\citep{egovjp}. Each Japanese governmental agency publishes its own annual white paper spanning a broad range of policy domains. While these are primarily distributed as PDFs, accurately extracting figures and tables from PDFs remains challenging even for current OCR models~\citep{wei2025deepseekocr,cui2025paddleocrvl}. We therefore restrict our dataset to white papers that also provide HTML editions, from which chart and table images can be collected directly via their URLs. Since different yearly editions of the same white paper often contain highly similar charts and tables (e.g., annual updates of population statistics), we use only the most recent edition from each white paper series. This choice also helps reduce potential contamination risk~\citep{oren2024proving}. As a result, our benchmark is constructed from 33 distinct governmental white paper series. The full list is provided in Appendix~\ref{sec:appendix_whitepapers}.

\paragraph{Image filtering and candidate selection.}
From these white papers, we initially collect 18{,}539 images. However, many of them are unsuitable for chart and table QA annotation, including photographs such as group pictures, low-resolution images whose contents are difficult to read, and near-duplicate images appearing across multiple pages. We therefore manually remove such inappropriate images and retain 5{,}903 candidate chart and table images for annotation.

\subsection{QA Annotation}
\label{subsec:qa_annotation}
Building on the collected chart and table images, we annotate them with question-answer pairs as follows.
QA annotation is carried out by 21 native Japanese-speaking annotators hired through a professional annotation agency.
Each annotator is given a set of images and writes one QA pair per image, following the requirements below.

\paragraph{QA requirements.}
Questions are required to
(a) be unanswerable from the question text alone, requiring the image to resolve;
(b) be natural and well-posed; and
(c) admit a single, unambiguous answer expressible in a word, phrase,
    or short sentence.
These constraints follow prior work showing that ambiguous questions make evaluation difficult and hinder fair assessment of model capabilities~\citep{chen2024mmstar,joshi2026datbench}.

We adopt a short answer format~\citep{wang2024charxiv}, avoiding multiple-choice and yes/no questions, which can be solved without the image or by random guessing~\citep{chen2024mmstar}. Answers are scored using an LLM-based judge that classifies each response as correct or incorrect, enabling robust evaluation that accommodates surface-level variations (e.g., ``7'' vs. ``seven'').

\paragraph{Question-type flags.}
To encourage diverse and reasoning-intensive QA pairs, each accepted question must satisfy at least one of the following question types:
\begin{itemize}[noitemsep]
\item \textbf{Global}: requiring integration of information spread across multiple regions of the image;
\item \textbf{Multi-hop}: requiring multiple reasoning steps, such as extracting multiple values and then computing their ratio;
\item \textbf{Counting}: requiring counting or indexing of objects;
\item \textbf{External knowledge}: requiring knowledge beyond the image to answer correctly, such as knowledge of Japanese geography;
\item \textbf{Visual}: requiring fine-grained visual perception, such as object color and shape;
\item \textbf{Other}: covering difficult questions that did not fit the above categories.
\end{itemize}
Annotators assign one or more flags per QA pair. The flags are non-exclusive by design: a single question frequently carries several (e.g., extracting values from distant regions of a table and then computing their ratio is both Global and Multi-hop).

Annotators are allowed to skip images for which no sufficiently challenging and well-defined question could be written, preventing the inclusion of artificially simple QA pairs.

\subsection{QA Verification}
\label{subsec:qa_verification}

QA annotation is prone to producing ambiguous questions and incorrectly annotated answers, and verification by independent annotators is widely used to improve benchmark quality~\citep{chen2024mmstar,masry2025chartqapro}. Following this practice, we introduce a separate verification stage for all annotated QA pairs.

For each QA pair, a second annotator (different from the original author) is given only the image and question and asked to independently answer it without access to the original answer.
If the two answers match up to minor surface variations, the example is accepted; otherwise, the QA pair is revised or discarded.
After completing the full pipeline, we obtained 2{,}053 QA pairs.

\begin{figure}[t]
  \centering
  \includegraphics[width=\linewidth]{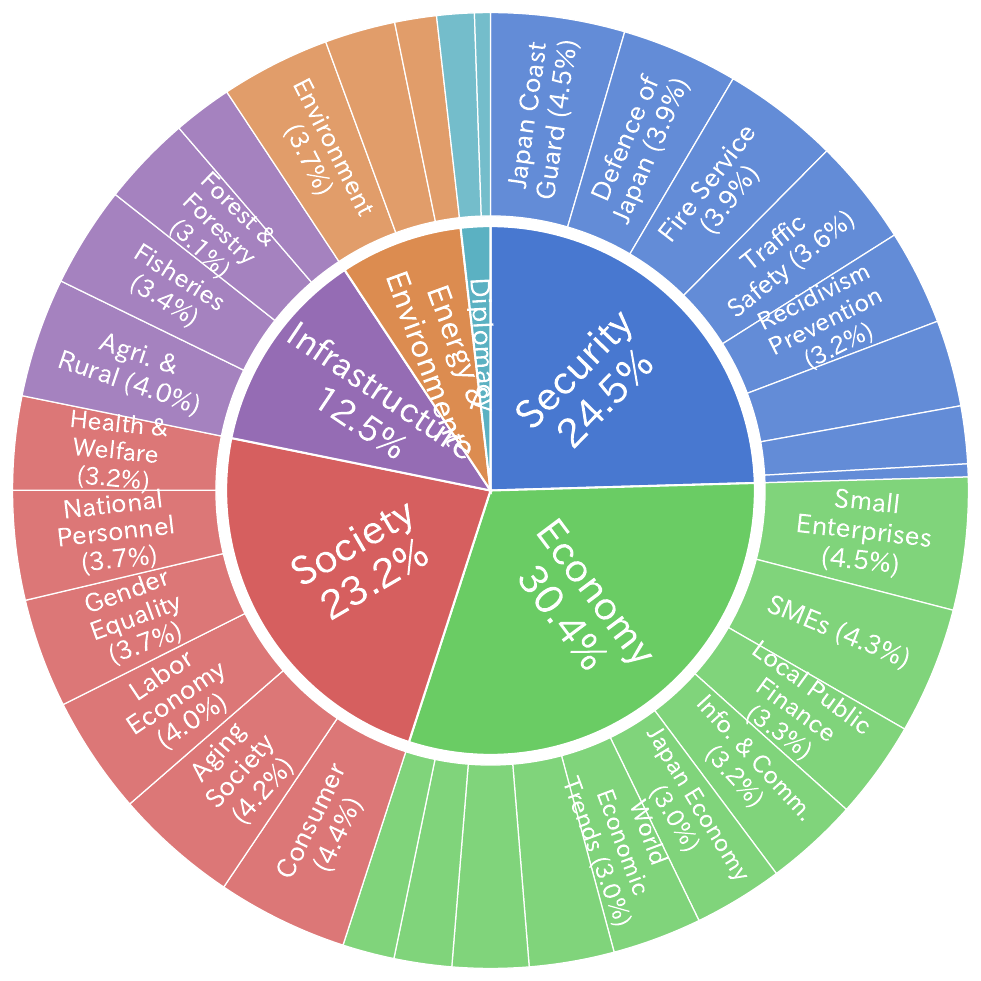}
  \caption{Distribution of QA pairs in \method{}. The inner ring represents the topic-wise distribution of examples, while the outer ring represents the white-paper-wise distribution.}
  \label{fig:topic_distribution}
\end{figure}

\begin{table*}[t]
\centering
\small
\setlength{\tabcolsep}{3pt}
\begin{tabular}{lrrrrrrrrrrr r}
\toprule
\textbf{} & \textbf{Bar} & \textbf{Line} & \textbf{Pie} & \textbf{Area} & \textbf{Scatter} & \textbf{Bubble} & \textbf{Map} & \textbf{Table} & \textbf{Infographic} & \textbf{Dashboard} & \textbf{Other} & \textbf{Total}\\
\midrule
\method{}   & 550   & 281 & 103 & 9  & 9  & 8  & 59 & 422 & 310 & 279 & 23  & \textbf{2{,}053}\\
JGraphQA    & 50    & 30  & 44  & 0  & 0  & 0  & 0  & 50  & 0   & 22  & 0   & 196\\
ChartQA     & 1{,}220 & 211 & 78 & 0 & 0  & 0  & 0  & 0   & 0   & 0   & 0   & 1{,}509\\
ChartQAPro  & 380   & 317 & 22  & 39 & 18 & 12 & 9  & 1   & 33  & 416 & 5   & 1{,}252\\
CharXiv     & 108   & 443 & 5   & 18 & 126 & 6 & 9  & 5   & 1   & 105 & 153 & 979\\
\bottomrule
\end{tabular}
\caption{Distribution of image types across benchmarks. Compared with JGraphQA, \method{} contains over ten times more examples and covers a broader range of image types, while remaining competitive with benchmarks such as ChartQAPro and CharXiv in both scale and diversity.}
\label{tab:image_types_full}
\end{table*}

\begin{table}[t]
\centering
\small
\begin{tabular}{lr}
\toprule
\textbf{Question Type} & \textbf{Examples}\\
\midrule
Global & 1{,}743\\
Multi-hop & 634\\
Counting & 293\\
External knowledge & 148\\
Visual & 897\\
Other & 132\\
\bottomrule
\end{tabular}
\caption{Distribution of question-type flags over the 2{,}053 verified QA pairs in \method{}. Each QA pair may have multiple flags, so counts are not mutually exclusive and do not sum to 2{,}053.}
\label{tab:question_type}
\end{table}

\section{Exploring \method{}}
\label{sec:exploring}

\paragraph{Statistics of \method{}.}
\method{} contains 2{,}053 examples collected from 33 white papers, grouped into six topics (Security, Economy, Society, Infrastructure, Energy \& Environment, and Diplomacy). Figure~\ref{fig:topic_distribution} shows the distribution of QA pairs per topic and per white paper, and Appendix~\ref{sec:appendix_whitepapers} lists the per-white-paper breakdown. The largest topic is Economy (30.4\%), followed by Society (23.3\%), while the number of QA pairs per white paper is relatively balanced.
Table~\ref{tab:question_type} reports the distribution of question-type flags. Each type has at least 100 examples, and Global is the most frequent, as multiple flags are allowed per question and Global tends to overlap with other types: 1{,}430 questions (69.7\%) carry two or more flags, most often Global together with Visual (801) or Multi-hop (604). Appendix~\ref{sec:appendix_question_types} gives the full co-occurrence matrix and one example per type. Questions are short (mean 39.9 characters, median 37, range 10--109), and answers are shorter still (median 5 characters).

\paragraph{Visual diversity.}

To analyze the visual diversity of \method{}, we categorize images from \method{}, JGraphQA, ChartQA (test), ChartQAPro, and CharXiv (val) into 11 visual-format categories: Bar, Line, Pie, Area, Scatter, Bubble, Map, Table, Infographic, Dashboard, and Other. Our taxonomy follows that of ChartQAPro~\citep{masry2025chartqapro}, with the addition of Map and Table. We use Gemini~3~Pro~\citep{google2025gemini3pro} to classify image types by providing each image together with a classification prompt, which is described in Appendix~\ref{sec:prompt}. To validate the classifier, an author blindly re-annotated 100 randomly sampled images without seeing the model's labels; the two agree on 83, with near-perfect agreement on single-chart types and disagreements concentrated on composite images for which a single label is inherently ambiguous (Appendix~\ref{sec:appendix_classifier}).
Table~\ref{tab:image_types_full} shows the image types distribution for each benchmark.
Compared to JGraphQA, \method{} contains more than 10$\times$ as many images and covers more than 10 image types, including types absent from JGraphQA such as Map and Infographic, enabling evaluation of more diverse visual understanding capabilities.

We further measure visual diversity using image embeddings extracted by the SigLIP2 vision encoder (\texttt{siglip2-so400m-patch16-512})~\citep{tschannen2025siglip2}. Computing the mean pairwise cosine distance between embeddings shows that \method{} achieves higher diversity than JGraphQA ($0.365$ vs.\ $0.275$).

\begin{figure*}[t]
  \includegraphics[width=\linewidth]{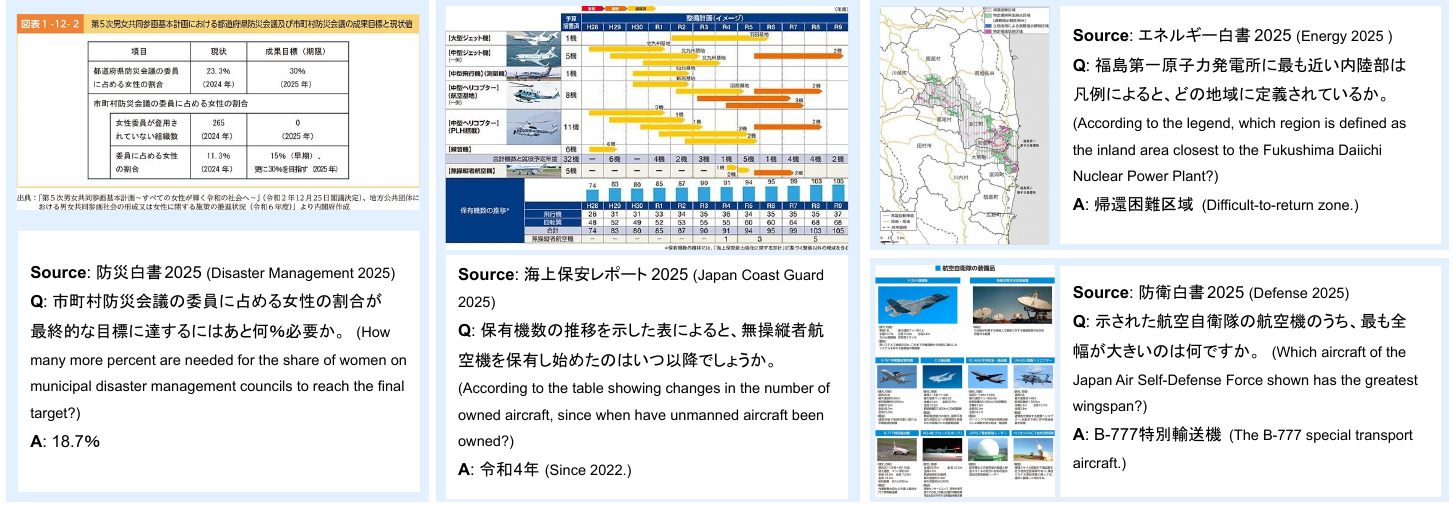}
  \caption{Representative VQA pairs in \method{}, requiring multi-hop reasoning and global image understanding rather than local visual cues alone.}
  \label{fig:hakushobench_examples}
\end{figure*}

\paragraph{Dataset showcase.}
Figure~\ref{fig:showcase_by_type} presents a randomly selected image from each image type in \method{}, illustrating its diversity. As shown, \method{} contains a wide variety of chart and table image types spanning diverse domains, reflecting the breadth of governmental white papers as a benchmark resource.
Figure~\ref{fig:hakushobench_examples} shows representative VQA pairs, demonstrating that \method{} contains challenging questions over information-dense chart and table images. For example, some questions require extracting and computing multiple numerical values from a table, others demand understanding spatial relationships between map locations and legends, and still others involve reading and comparing textual descriptions embedded in figures. These examples illustrate that \method{} often requires multi-hop reasoning and holistic image understanding beyond simple value extraction.

\begin{figure*}[t]
  \includegraphics[width=\linewidth]{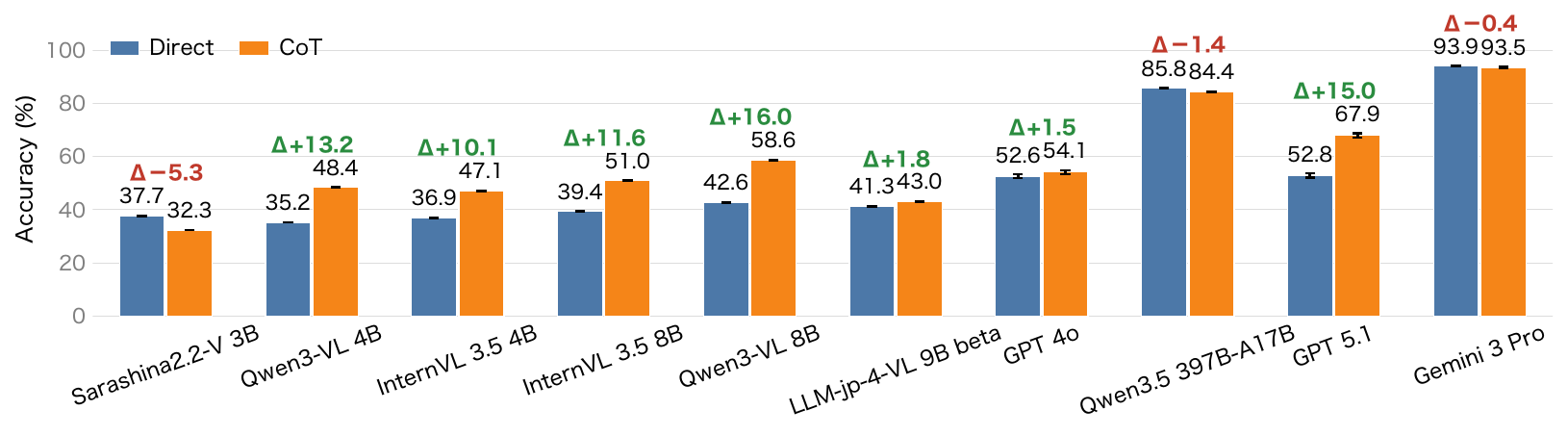}
  \caption{Performance of each model on \method{} under the Direct and CoT settings. Gemini~3~Pro and Qwen3.5-397B-A17B reason internally in both settings, so their Direct bars already include native thinking. All sub-10B open-weight models remain below 60\% accuracy, highlighting the challenging nature of \method{}.}
  \label{fig:hakushobench_main_result}
\end{figure*}

\section{Experiments}
\label{sec:experiments}

We evaluate a broad set of open-weight and proprietary models on \method{}, and compare its characteristics against existing benchmarks.

\subsection{Experimental Settings}
\label{subsec:exp_settings}

\paragraph{Models.}
We evaluate a diverse set of open-weight and proprietary VLMs. On the open-weight side, we include Qwen3-VL-{4B, 8B}~\citep{bai2025qwen3vl} and InternVL3.5-{4B, 8B}~\citep{wang2025internvl35} as general-purpose multilingual models, as well as Sarashina2.2-Vision-3B~\citep{sbintuitions2025sarashina} and LLM-jp-4-VL~9B~beta~\citep{sugiura2026jagle} as Japanese-centric models. To gauge the frontier of open-weight models, we additionally evaluate Qwen3.5-397B-A17B~\citep{qwen35blog}, a 397B-parameter mixture-of-experts VLM with 17B active parameters. On the proprietary side, we evaluate GPT-4o (\texttt{gpt-4o-2024-11-20})~\citep{openai2024gpt4ocard}, GPT-5.1 (\texttt{gpt-5.1-2025-11-13})~\citep{openai2025gpt5.1}, and Gemini~3~Pro (\texttt{gemini-3-pro-preview})~\citep{google2025gemini3pro}.

\paragraph{Prompt settings.}
To examine the effect of reasoning-oriented prompting, we evaluate two settings per model. The \textbf{Direct} setting asks models to answer the question accurately and concisely, while the \textbf{CoT} setting additionally instructs models to think step by step before producing the final answer~\citep{wei2022chain,kojima2022large}. Full prompts are provided in Appendix~\ref{sec:prompt}.

\paragraph{Inference settings.}
We set the temperature to 0 for all models except GPT-5.1, which does not support temperature control, and use a maximum generation length of 8{,}192 tokens. Open-weight models are evaluated on NVIDIA A100 GPUs, except Qwen3.5-397B-A17B, which is served with SGLang on 8 NVIDIA H200 GPUs with its native thinking mode enabled and the reasoning trace stripped before grading. Proprietary models are accessed through their official APIs. For reasoning settings, we use the \texttt{medium} reasoning mode for Gemini~3~Pro. For GPT-5.1, we use \texttt{none} for the Direct setting and \texttt{medium} for the CoT setting.

\paragraph{Evaluation metric.}
We report accuracy, scoring each model output as correct or incorrect with GPT-5.1 (\texttt{gpt-5.1-2025-11-13}) as an LLM judge, which tolerates minor surface-level variations in phrasing. To account for stochastic variation in both model outputs and LLM judgment, each evaluation is repeated three times, and we report the mean and standard deviation. The judge prompt is given in Appendix~\ref{sec:prompt}, and Appendix~\ref{sec:appendix_judge} validates the judge against human judgment (95\% agreement, Cohen's $\kappa=0.90$).

\paragraph{Compared benchmarks.}
To better understand the characteristics of \method{}, we additionally evaluate the same set of models on four existing chart and table benchmarks: JGraphQA~\citep{jgraphqa}, ChartQA (test)~\citep{masry2022chartqa}, ChartQAPro~\citep{masry2025chartqapro}, and CharXiv (val)~\citep{wang2024charxiv}. For JGraphQA, we use JGraphQA-Verified~\citep{sugiura2026jammeval}, a cleaned and corrected version of the original benchmark. Per-model results on these benchmarks are visualized in Appendix~\ref{sec:other_benchmark_results}, and representative images from each benchmark are shown in Appendix~\ref{sec:appendix_showcase_other}.

\section{Results}
\label{sec:results}

\begin{table*}[t]
\centering
\small
\setlength{\tabcolsep}{4pt}
\resizebox{\textwidth}{!}{%
\begin{tabular}{lcccccccccc}
\toprule
\textbf{Model} & \multicolumn{2}{c}{\textbf{\method{}}} & \multicolumn{2}{c}{\textbf{JGraphQA}} & \multicolumn{2}{c}{\textbf{ChartQA}} & \multicolumn{2}{c}{\textbf{ChartQAPro}} & \multicolumn{2}{c}{\textbf{CharXiv}} \\
\cmidrule(lr){2-3} \cmidrule(lr){4-5} \cmidrule(lr){6-7} \cmidrule(lr){8-9} \cmidrule(lr){10-11}
 & Direct & CoT & Direct & CoT & Direct & CoT & Direct & CoT & Direct & CoT \\
\midrule
\rowcolor{gray!10}
\textit{Open-weight Models} &&&&&&&&&&\\
Sarashina2.2-V 3B & 37.7{\tiny\,$\pm$\,0.1} & 32.3{\tiny\,$\pm$\,0.1} & 80.4{\tiny\,$\pm$\,0.3} & 81.0{\tiny\,$\pm$\,0.8} & 47.8{\tiny\,$\pm$\,0.2} & 63.5{\tiny\,$\pm$\,0.1} & 23.0{\tiny\,$\pm$\,0.0} & 23.0{\tiny\,$\pm$\,0.1} & 48.6{\tiny\,$\pm$\,0.0} & 50.7{\tiny\,$\pm$\,0.0} \\
Qwen3-VL 4B & 35.2{\tiny\,$\pm$\,0.1} & 48.4{\tiny\,$\pm$\,0.1} & 84.2{\tiny\,$\pm$\,0.5} & 91.5{\tiny\,$\pm$\,0.3} & 67.0{\tiny\,$\pm$\,0.1} & 83.7{\tiny\,$\pm$\,0.0} & 30.5{\tiny\,$\pm$\,0.0} & 30.5{\tiny\,$\pm$\,0.1} & 65.3{\tiny\,$\pm$\,0.0} & 68.3{\tiny\,$\pm$\,0.1} \\
InternVL 3.5 4B & 36.9{\tiny\,$\pm$\,0.1} & 47.1{\tiny\,$\pm$\,0.2} & 84.2{\tiny\,$\pm$\,0.0} & 84.2{\tiny\,$\pm$\,0.0} & 73.7{\tiny\,$\pm$\,0.1} & 82.1{\tiny\,$\pm$\,0.1} & 34.7{\tiny\,$\pm$\,0.1} & 34.8{\tiny\,$\pm$\,0.1} & 62.5{\tiny\,$\pm$\,0.0} & 61.9{\tiny\,$\pm$\,0.0} \\
InternVL 3.5 8B & 39.4{\tiny\,$\pm$\,0.0} & 51.0{\tiny\,$\pm$\,0.0} & 85.9{\tiny\,$\pm$\,0.3} & 88.4{\tiny\,$\pm$\,0.3} & 75.0{\tiny\,$\pm$\,0.2} & 81.1{\tiny\,$\pm$\,0.3} & 33.5{\tiny\,$\pm$\,0.0} & 33.5{\tiny\,$\pm$\,0.1} & 66.8{\tiny\,$\pm$\,0.0} & 65.7{\tiny\,$\pm$\,0.1} \\
Qwen3-VL 8B & 42.6{\tiny\,$\pm$\,0.2} & 58.6{\tiny\,$\pm$\,0.2} & 86.6{\tiny\,$\pm$\,0.3} & 88.8{\tiny\,$\pm$\,0.5} & 70.3{\tiny\,$\pm$\,0.0} & 84.3{\tiny\,$\pm$\,0.3} & 35.0{\tiny\,$\pm$\,0.1} & 35.1{\tiny\,$\pm$\,0.1} & 70.1{\tiny\,$\pm$\,0.1} & 71.2{\tiny\,$\pm$\,0.1} \\
LLM-jp-4-VL 9B beta & 41.3{\tiny\,$\pm$\,0.1} & 43.0{\tiny\,$\pm$\,0.1} & 87.8{\tiny\,$\pm$\,0.5} & 85.2{\tiny\,$\pm$\,0.0} & 70.5{\tiny\,$\pm$\,0.1} & 68.0{\tiny\,$\pm$\,0.1} & 27.8{\tiny\,$\pm$\,0.2} & 27.8{\tiny\,$\pm$\,0.1} & 58.4{\tiny\,$\pm$\,0.1} & 56.6{\tiny\,$\pm$\,0.1} \\
Qwen3.5 397B-A17B & 85.8{\tiny\,$\pm$\,0.1} & 84.4{\tiny\,$\pm$\,0.2} & 98.0{\tiny\,$\pm$\,0.5} & \textbf{98.1{\tiny\,$\pm$\,0.3}} & \textbf{86.1{\tiny\,$\pm$\,0.3}} & 79.6{\tiny\,$\pm$\,0.1} & 60.5{\tiny\,$\pm$\,0.2} & 60.7{\tiny\,$\pm$\,0.3} & 79.3{\tiny\,$\pm$\,0.1} & 79.0{\tiny\,$\pm$\,0.0} \\
\midrule
\rowcolor{gray!10}
\textit{Proprietary Models} &&&&&&&&&&\\
GPT 4o & 52.6{\tiny\,$\pm$\,0.6} & 54.1{\tiny\,$\pm$\,0.7} & 92.0{\tiny\,$\pm$\,0.6} & 84.7{\tiny\,$\pm$\,1.0} & 62.1{\tiny\,$\pm$\,0.3} & 67.1{\tiny\,$\pm$\,0.3} & 44.1{\tiny\,$\pm$\,0.2} & 48.4{\tiny\,$\pm$\,0.1} & 71.6{\tiny\,$\pm$\,0.1} & 71.4{\tiny\,$\pm$\,0.2} \\
GPT 5.1 & 52.8{\tiny\,$\pm$\,0.8} & 67.9{\tiny\,$\pm$\,0.8} & 95.1{\tiny\,$\pm$\,1.1} & 94.9{\tiny\,$\pm$\,1.0} & 64.9{\tiny\,$\pm$\,0.7} & 77.4{\tiny\,$\pm$\,0.3} & 42.0{\tiny\,$\pm$\,0.4} & 55.5{\tiny\,$\pm$\,0.7} & 68.9{\tiny\,$\pm$\,0.2} & 75.8{\tiny\,$\pm$\,0.1} \\
Gemini 3 Pro & \textbf{93.9{\tiny\,$\pm$\,0.1}} & 93.5{\tiny\,$\pm$\,0.2} & 97.3{\tiny\,$\pm$\,0.3} & 96.9{\tiny\,$\pm$\,0.5} & 80.5{\tiny\,$\pm$\,0.4} & 85.7{\tiny\,$\pm$\,0.4} & \textbf{67.4{\tiny\,$\pm$\,0.2}} & 65.8{\tiny\,$\pm$\,0.2} & 81.4{\tiny\,$\pm$\,0.2} & \textbf{81.6{\tiny\,$\pm$\,0.1}} \\
\bottomrule
\end{tabular}
}
\caption{Performance of each model on chart and table benchmarks under the Direct and CoT settings, reported as accuracy (\%) averaged over three runs (mean $\pm$ standard deviation). For each benchmark, the best score across all (model, setting) pairs is shown in bold. Gemini~3~Pro and Qwen3.5-397B-A17B reason internally in both settings, so their Direct columns already include native thinking.}
\label{tab:full_results}
\end{table*}

\subsection{Main Results}
\label{subsec:main_results}
Figure~\ref{fig:hakushobench_main_result} shows model performance on \method{}.

\paragraph{Gemini~3~Pro leads; the flagship open-weight model narrows but does not close the gap.}
Gemini~3~Pro achieves the highest score of $93.9\%$. Among open-weight models, Qwen3.5-397B-A17B reaches $85.8\%$, clearly surpassing GPT-5.1 ($67.9\%$) but still 8.1 points below Gemini~3~Pro. All sub-10B open-weight models remain below $60\%$, with Qwen3-VL~8B the best at $58.6\%$. GPT-5.1 scores only $67.9\%$, revealing a substantial gap even among proprietary models in Japanese chart and table understanding. These results suggest considerable room for improvement in open-weight models, and in particular in the small models that are most widely deployed.

\paragraph{CoT prompting effectiveness varies by model.}
CoT prompting improves accuracy for most models, with GPT~5.1, Qwen3-VL, and InternVL3.5 all gaining more than 10 points. In contrast, GPT-4o, LLM-jp-4-VL~9B~beta, and Sarashina2.2-Vision~3B show limited gains or even degradation. Manual analysis suggests these models often fail to engage in reasoning or produce repetitive and incoherent chains, indicating limited reasoning ability. The two models with native thinking, Gemini~3~Pro and Qwen3.5-397B-A17B, already reason internally in the Direct setting and gain nothing from the additional CoT prompt ($-0.4$ and $-1.4$ points).

\begin{figure*}[t]
  \centering
  \includegraphics[width=\linewidth]{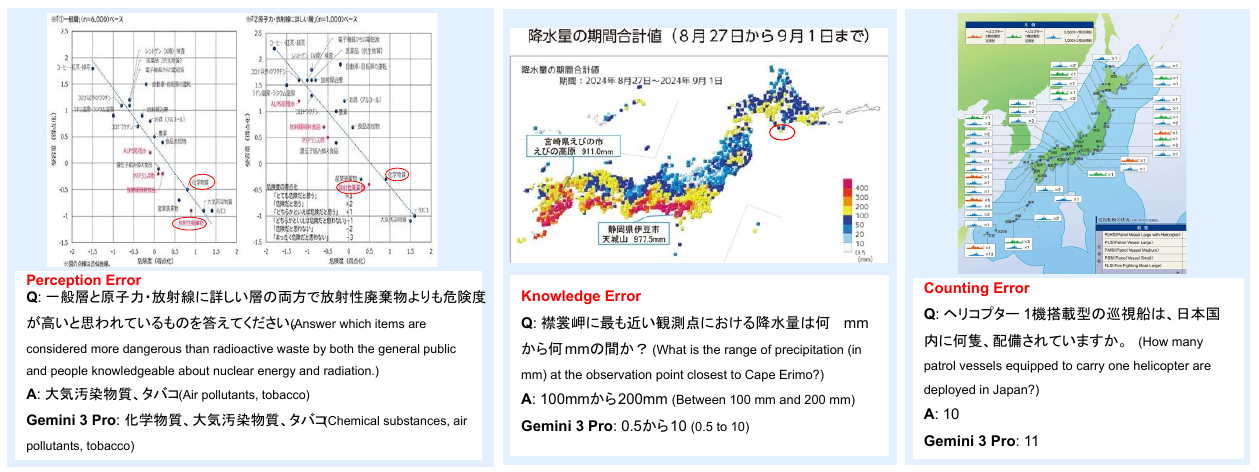}
  \caption{Representative failure cases of Gemini~3~Pro on \method{}. Left: a perception error in reading scatter plot conditions. Middle: a knowledge error in identifying the map location closest to Cape Erimo (襟裳岬). Right: a counting error resulting in an off-by-one prediction.}
  \label{fig:gemini3pro_errors}
\end{figure*}

\subsection{Comparison with Existing Benchmarks}
\label{subsec:bench_comparison}
Table~\ref{tab:full_results} reports the performance of each model on all benchmarks, and Figure~\ref{fig:bench_range} summarizes the score range across models.
\paragraph{\method{} is more challenging than JGraphQA.}
Compared to JGraphQA, \method{} is much harder for every model: Qwen3-VL~8B reaches 58.6\% on \method{} versus 88.8\% on JGraphQA, Sarashina2.2-Vision~3B 37.7\% versus 81.0\%, and even Qwen3.5-397B-A17B 85.8\% versus 98.1\%.
Appendix~\ref{sec:appendix_question_type} shows the accuracy spread across image types and across question types. The spread is consistent across both dimensions, with infographics being a notable exception, suggesting that \method{} evaluates a broad range of model capabilities.

\paragraph{\method{} exposes performance gaps that saturated benchmarks hide.}
On JGraphQA and ChartQA, Qwen3.5-397B-A17B matches or exceeds Gemini~3~Pro (98.1 vs.\ 97.3 and 86.1 vs.\ 85.7), so these benchmarks no longer separate the two, whereas on \method{} an 8.1-point gap remains. The same holds across the model range: the spread between the weakest and strongest model is 56.3 points on \method{} (37.7--93.9), compared with 17.2 on JGraphQA (81.0--98.1) and 22.6 on ChartQA (63.5--86.1) (Figure~\ref{fig:bench_range}). In this respect \method{} behaves like the recent English benchmarks designed to remain challenging, CharXiv (30.9) and ChartQAPro (44.4), which likewise keep models clearly distinguishable.

\subsection{Does \method{} Require Visual Understanding?}
\label{subsec:input_ablation}

\begin{table}[t]
\centering
\begin{tabular*}{\columnwidth}{@{\extracolsep{\fill}}lc@{}}
\toprule
\textbf{Visual input} & \textbf{Accuracy (\%)} \\
\midrule
Original image (default)      & 42.6{\small\,$\pm$\,0.2} \\
OCR text of the image         & 28.2{\small\,$\pm$\,0.1} \\
None (question only)          & \phantom{0}6.1{\small\,$\pm$\,0.0} \\
\bottomrule
\end{tabular*}
\caption{Input ablation of Qwen3-VL~8B on \method{} (Direct setting). The question is always given; only the visual input varies. OCR text is produced by PaddleOCR-VL-1.6.}
\label{tab:input_ablation}
\end{table}

To verify that \method{} requires understanding the visual content of the image, not only the text it contains, we evaluate Qwen3-VL~8B in the Direct setting under three conditions that keep the question fixed and vary only the visual input (Table~\ref{tab:input_ablation}): the original image, the image replaced by its OCR text, and no visual input at all. OCR text is produced by PaddleOCR-VL-1.6~\citep{zhang2026paddleocrvl16}, a state-of-the-art document parsing model.

Replacing the image with its OCR text drops accuracy from 42.6\% to 28.2\%, so text extraction alone is not sufficient and \method{} requires visual and layout understanding beyond the textual content of the image. Removing the visual input entirely drops accuracy to 6.1\%, so \method{} cannot be solved from textual priors or memorized statistics.

\subsection{Error Analysis of Gemini~3~Pro}
\label{subsec:error_analysis}

To better understand the remaining challenges in \method{}, we manually analyze 50 randomly sampled questions incorrectly answered by Gemini~3~Pro (CoT setting), the best-performing model on the benchmark.
Of the 50 cases, 22 are perception errors, 3 are reasoning errors, and one each are OCR, external knowledge, and counting errors; the remaining 22 are attributable to gold-answer ambiguity (13) or judge false negatives (9), so the true error rate of Gemini~3~Pro is lower than the reported 6--7\%.
Figure~\ref{fig:gemini3pro_errors} shows representative examples of three error types.
In the \emph{Perception} example, the model fails to correctly interpret spatial relationships in a scatter plot. The \emph{Knowledge} example requires external geographic knowledge to identify Cape Erimo on a thematic precipitation map. The \emph{Counting} example illustrates an off-by-one error in fine-grained counting.
Even the best-performing model thus still makes diverse errors, dominated by fine-grained perception.

\section{Conclusion}
\label{sec:conclusion}
We presented \method{}, a Japanese chart and table VQA benchmark constructed from 33 governmental white papers, containing 2,053 VQA pairs over 10 distinct image types. We demonstrated that governmental white papers serve as a valuable source for benchmark construction, offering broad domain coverage and visual diversity. \method{} is substantially harder than JGraphQA and remains challenging for open-weight models: sub-10B models reach at most 58.6\% accuracy, and even the flagship Qwen3.5-397B-A17B trails Gemini~3~Pro by 8.1 points despite matching it on JGraphQA and ChartQA.

\section*{Limitations}

\paragraph{Language and domain coverage.}
\method{} addresses the lack of challenging and visually diverse Japanese chart and table QA benchmarks. However, the dataset is currently limited to Japanese and does not directly address the scarcity of benchmarks for other low-resource languages. In addition, our dataset is constructed exclusively from governmental white papers, which may not fully cover the diversity of visual styles and domains found in other real-world documents. Because many countries publish analogous governmental reports and white papers~\citep{usgov2026economic}, the same source-level methodology could be applied to other languages, but we have demonstrated it only for Japanese and leave multilingual instantiation to future work.

\paragraph{Potential data contamination.}
We use the most recent edition of each white paper to mitigate contamination risk~\citep{oren2024proving}. However, we cannot completely rule out the possibility that some images or related information appeared in derivative web content included in model pretraining corpora. 
That said, all QA pairs in our benchmark were newly constructed through manual annotation, making it highly unlikely that they were included in any model's pretraining data.

\paragraph{Saturation at the frontier.}
While \method{} proves challenging for open-weight models, Gemini~3~Pro achieves $93.9\%$, leaving limited headroom to discriminate among frontier models. A natural remedy is to construct a harder subset by filtering out questions solved by all evaluated models, following~\citet{phan2026hle}, or to collect more demanding QA pairs targeting capabilities beyond current frontier models. However, increasing difficulty introduces a risk of producing unnatural questions that deviate from realistic use cases. Balancing difficulty and practical relevance is therefore a non-trivial design challenge that we leave for future work.

\section*{Ethical Considerations}

\paragraph{Public data sources and safety.}
All images are collected from white papers publicly released by Japanese governmental agencies. Given this source, risks related to privacy, personally identifiable information, and NSFW (Not Safe For Work) content are negligible. This was further confirmed during manual filtering, where no problematic content was observed.

\section*{Acknowledgments}
In this research work, we used the ``mdx: a platform for building data-empowered society''.
We used ABCI 3.0 provided by AIST and AIST Solutions with support from ``ABCI 3.0
Development Acceleration Use''.
This research was supported by the Science Tokyo Support Program for Doctoral Students, funded by the Universities for International Research Excellence.

\bibliography{custom}

\appendix

\section{Licenses for Our Resources}
HakushoBench and its evaluation code are released under the Apache 2.0 License. Note that we distribute only image URLs rather than the raw image data.

\section{Use of AI Assistants}
We used AI assistants to correct typographical errors, improve the clarity and naturalness of expressions, and generate scripts for plotting figures.

\section{White Paper Sources}
\label{sec:appendix_whitepapers}

Table~\ref{tab:whitepaper_sources} lists the 33 Japanese governmental white papers used in \method{}.

\begin{table*}[t]
\centering
\small
\setlength{\tabcolsep}{2pt}
\begin{tabular}{lllr}
\toprule
\textbf{Japanese Title} & \textbf{English Title} & \textbf{Topic Group} & \textbf{Examples}\\
\midrule
海上保安レポート2025 & Japan Coast Guard 2025 & Security & 93\\
防衛白書2025 & Defence of Japan 2025 & Security & 81\\
消防白書2024 & Fire Service 2024 & Security & 81\\
交通安全白書2025 & Traffic Safety 2025 & Security & 74\\
再犯防止推進白書2024 & Recidivism Prevention 2024 & Security & 66\\
防災白書2025 & Disaster Management 2025 & Security & 60\\
犯罪被害者白書2024 & Crime Victims 2024 & Security & 40\\
警察白書2025 & Police 2025 & Security & 9\\
\midrule
小規模企業白書2025 & Small Enterprises 2025 & Economy & 92\\
中小企業白書2025 & Small and Medium Enterprises 2025 & Economy & 88\\
地方財政白書2025 & Local Public Finance 2025 & Economy & 68\\
情報通信白書2025 & Information \& Communications 2025 & Economy & 65\\
日本経済レポート2024 & Japan Economy 2024 & Economy & 62\\
世界経済の潮流2025-01 & World Economic Trends 2025 & Economy & 62\\
通商白書2025 & International Economy and Trade 2025 & Economy & 59\\
年次経済財政報告2025 & Japanese Economy and Public Finance 2025 & Economy & 53\\
地域課題分析レポート2025-08 & Regional Issues 2025 & Economy & 40\\
科学技術白書2025 & Science \& Technology 2025 & Economy & 36\\
\midrule
消費者白書2023 & Consumer 2023 & Society & 91\\
高齢社会白書2025 & Aging Society 2025 & Society & 86\\
労働経済白書2024 & Labor Economy 2024 & Society & 82\\
男女共同参画白書2025 & Gender Equality 2025 & Society & 76\\
人事院白書2024 & National Personnel 2024 & Society & 76\\
厚生労働白書2025 & Health, Labour \& Welfare 2025 & Society & 65\\
\midrule
食料農業農村白書2024 & Food, Agriculture and Rural Areas 2024 & Infrastructure & 83\\
水産白書2024 & Developments in Japan's Fisheries 2024 & Infrastructure & 70\\
森林林業白書2024 & Forest \& Forestry 2024 & Infrastructure & 63\\
国土交通白書2024 & Land, Infrastructure, Transport and Tourism 2024 & Infrastructure & 41\\
\midrule
環境白書2025 & Environment 2025 & Energy \& Environment & 76\\
原子力白書2023 & Nuclear Energy 2023 & Energy \& Environment & 49\\
エネルギー白書2025 & Energy 2025 & Energy \& Environment & 29\\
\midrule
開発協力白書2023 & Development Cooperation 2023 & Diplomacy & 26\\
外交青書2025 & Diplomatic Bluebook 2025 & Diplomacy & 11\\
\midrule
\textbf{Total} & & & 2{,}053\\
\bottomrule
\end{tabular}
\caption{The 33 Japanese white papers used in \method{}.}
\label{tab:whitepaper_sources}
\end{table*}

\section{Validation of the Image-Type Classifier}
\label{sec:appendix_classifier}

To validate the Gemini~3~Pro image-type labels used in Section~\ref{sec:exploring}, an author re-annotated 100 randomly sampled \method{} images without seeing the model's labels. Table~\ref{tab:classifier_confusion} shows the confusion matrix. The two agree on 83 of 100 images. Agreement on single-chart types is near-perfect (Table 19/19, Infographic 9/9, Map 3/3, Pie 2/2, Bar 29/30, Line 15/16), and 15 of the 17 disagreements involve the Dashboard class: composite images with several panels or mixed chart types, where the human tends to label the dominant panel while the model assigns Dashboard, and no single ground-truth label exists. The labels are therefore not a strict taxonomy for composite figures, but they are sufficiently accurate for their purpose in this paper, namely comparing the image-type diversity of benchmarks.

\begin{table*}[t]
\centering
\small
\setlength{\tabcolsep}{3pt}
\begin{tabular}{lrrrrrrrr}
\toprule
\textbf{Model $\backslash$ Human} & \textbf{Bar} & \textbf{Dash.} & \textbf{Info.} & \textbf{Line} & \textbf{Map} & \textbf{Other} & \textbf{Pie} & \textbf{Table} \\
\midrule
Bar         & \textbf{29} & 1 & 0 & 0 & 0 & 0 & 0 & 0 \\
Dashboard   & 9 & \textbf{6} & 1 & 1 & 2 & 1 & 0 & 0 \\
Infographic & 0 & 0 & \textbf{9} & 0 & 0 & 0 & 0 & 0 \\
Line        & 0 & 1 & 0 & \textbf{15} & 0 & 0 & 0 & 0 \\
Map         & 0 & 0 & 0 & 0 & \textbf{3} & 0 & 0 & 0 \\
Other       & 0 & 0 & 1 & 0 & 0 & \textbf{0} & 0 & 0 \\
Pie         & 0 & 0 & 0 & 0 & 0 & 0 & \textbf{2} & 0 \\
Table       & 0 & 0 & 0 & 0 & 0 & 0 & 0 & \textbf{19} \\
\bottomrule
\end{tabular}
\caption{Confusion matrix of image-type labels on 100 randomly sampled \method{} images (rows: Gemini~3~Pro; columns: human annotator). Overall agreement is 83/100.}
\label{tab:classifier_confusion}
\end{table*}

\section{Validation of the LLM Judge}
\label{sec:appendix_judge}

We validate the GPT-5.1 judge against human judgment on 100 model answers sampled uniformly at random from the outputs of all evaluated models and settings. An author graded each answer given the image, question, and gold answer. The judge and the human agree on 95 of 100 answers (Cohen's $\kappa = 0.90$). All five disagreements are in the same direction: the human judged the answer correct while the judge marked it incorrect, typically because of surface-form variation or because the answer is valid given the image content, which the text-only judge cannot see. There is no case in which the judge accepted an answer the human rejected, so the reported accuracies are, if anything, slightly conservative. Providing the image to the judge is a possible refinement that we leave for future work.

\section{Prompt}
\label{sec:prompt}
We show the prompts used in the experiments below.

\begin{Prompt}{Image type classification prompt}
You are an expert at classifying figures in documents.

Look at the image and choose exactly ONE label from the following 11
categories that best describes the primary visualization type:

- Bar         : Bar chart (vertical, horizontal, grouped, or stacked bars).

- Line        : Line chart (including multi-line time series).

- Pie         : Pie chart or donut chart.

- Area        : Area chart or stacked-area chart.

- Scatter     : Scatter plot.

- Bubble      : Bubble chart.

- Map         : Geographic / thematic / choropleth map.

- Table       : A pure table with no charts.

- Infographic : Text-heavy figure with icons, illustrations, or flow
                elements; not a chart or table.
                
- Dashboard   : A composite figure that combines two or more of the above
                in one image (e.g., a chart together with an inline table,
                or multiple charts of different types laid out together).
                
- Other       : Does not fit any of the above.

Rules:
- Choose EXACTLY ONE label.

- If the image contains multiple chart components (e.g., a bar chart and
  an inline table, or two different chart types side by side), use
  "Dashboard".
  
- Use "Other" only as a last resort.

Respond with ONLY the label (e.g., "Bar"). No explanation, no quotes.
\end{Prompt}

\begin{Prompt}{Direct prompt}
上記の質問に対して、正確かつ簡潔に答えてください。

(For the above question, answer accurately and concisely.)
\end{Prompt}

\begin{Prompt}{CoT prompt}
上記の質問に対して、ステップバイステップで考えてから答えてください。

最後の行は 'Answer: \$ANSWER' の形式で、正確かつ簡潔に回答してください。

(For the above question, think step by step before answering.

The final line should be in the format 'Answer: \$ANSWER', and provide the answer accurately and concisely.)
\end{Prompt}

\begin{Prompt}{Judge prompt}
Judge whether the following [response] to [question] is correct or not based on the precise and unambiguous [correct\_answer] below.

When judging equivalence, allow variations in script or notation that convey the same meaning
(e.g., '2羽' and '二羽' should be considered equivalent).

Treat the following cases as correct:
- The extracted answer includes additional context (e.g., series name, author name, location, broader category) while still containing the correct\_answer exactly or as its unambiguous, specific instance.
  (For example, "富嶽三十六景 江戸日本橋" is correct if the correct\_answer is "江戸日本橋".)
- The extracted answer is more specific than the [correct\_answer] while remaining consistent with it.
- The extracted answer is an alternate name, synonymous phrasing, or another commonly accepted way to refer to the same concept, object, place, or artwork.
- The extracted answer omits information that is not essential to the correctness of the question.
- Allow minor variations in spacing, capitalization, or script, as long as the core correct\_answer is unambiguously present.

[question]: {question}

[response]: {response}

Your judgement must be in the format and criteria specified below:

extracted\_final\_answer: The final exact answer extracted from the [response]. Put the extracted answer as 'None' if there is no exact, final answer to extract from the response.

[correct\_answer]: {correct\_answer}

reasoning: Explain why the extracted\_final\_answer is correct or incorrect based on [correct\_answer], focusing only on if there are meaningful differences between [correct\_answer] and the extracted\_final\_answer. Do not comment on any background to the problem, do not attempt to solve the problem, do not argue for any answer different than [correct\_answer], focus only on whether the answers match.

correct: Answer 'yes' if extracted\_final\_answer matches the [correct\_answer] given above, or is within a small margin of error for numerical problems. Answer 'no' otherwise, i.e. if there is any inconsistency, ambiguity, non-equivalency, or if the extracted answer is incorrect.

confidence: The extracted confidence score between 0\% and 100\% from [response]. Put 100 if there is no confidence score available.
\end{Prompt}

\section{Question Types: Co-occurrence and Examples}
\label{sec:appendix_question_types}

Question-type flags are non-exclusive. Table~\ref{tab:qtype_cooccurrence} reports how often each pair of flags is assigned to the same question, and Figure~\ref{fig:qtype_examples} shows one representative question per type.

\begin{table}[t]
\centering
\footnotesize
\setlength{\tabcolsep}{1.8pt}
\begin{tabular}{lrrrrrr}
\toprule
 & \textbf{Glob.} & \textbf{Multi.} & \textbf{Count.} & \textbf{Ext.} & \textbf{Vis.} & \textbf{Other} \\
\midrule
Global (Glob.)      & 1{,}743 & 604 & 231 & 110 & 801 & 0 \\
Multi-hop (Multi.)  &  & 634 & 40 & 39 & 206 & 0 \\
Counting (Count.)   &  &  & 293 & 13 & 64 & 0 \\
Ext.\ knowl.\ (Ext.) &  &  &  & 148 & 34 & 0 \\
Visual (Vis.)       &  &  &  &  & 897 & 0 \\
Other               &  &  &  &  &  & 132 \\
\bottomrule
\end{tabular}
\caption{Co-occurrence of question-type flags in \method{}. The diagonal gives the number of questions carrying each flag; off-diagonal cells give the number of questions carrying both flags. Other is assigned only when no other flag applies.}
\label{tab:qtype_cooccurrence}
\end{table}

\newcommand{\qtimg}[1]{\parbox[c][4.2cm][c]{\linewidth}{\centering\includegraphics[width=\linewidth,height=4.2cm,keepaspectratio]{images/qtype_examples/#1}}}
\newcommand{\qtcard}[1]{{\small\textbf{#1}}}
\begin{figure*}[t]
\centering
\begin{minipage}[t]{0.32\textwidth}
  \centering
  \qtcard{Global}\\[2pt]
  \qtimg{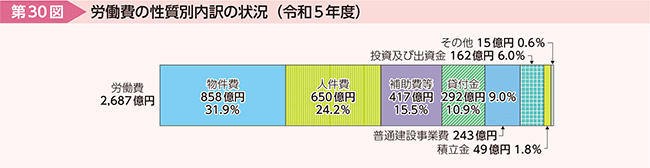}\\[3pt]
  \begin{minipage}{\linewidth}\footnotesize\raggedright
  \textbf{Q:} 令和5年度の労働費のうち、2番目に大きな金額の費目は何ですか。 \emph{(Among the FY2023 labor expenditure items, which item has the second-largest amount?)}\\
  \textbf{A:} 人件費 \emph{(Personnel expenses)}
  \end{minipage}
\end{minipage}\hfill
\begin{minipage}[t]{0.32\textwidth}
  \centering
  \qtcard{Multi-hop}\\[2pt]
  \qtimg{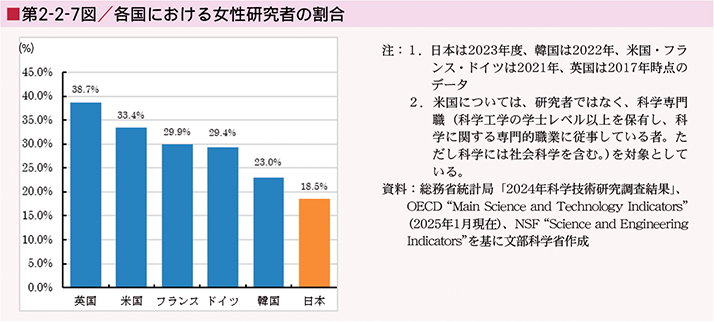}\\[3pt]
  \begin{minipage}{\linewidth}\footnotesize\raggedright
  \textbf{Q:} 女性研究者の割合が日本の2倍以上の国は？ \emph{(Which country has a share of female researchers more than twice that of Japan?)}\\
  \textbf{A:} 英国 \emph{(United Kingdom)}
  \end{minipage}
\end{minipage}\hfill
\begin{minipage}[t]{0.32\textwidth}
  \centering
  \qtcard{Counting}\\[2pt]
  \qtimg{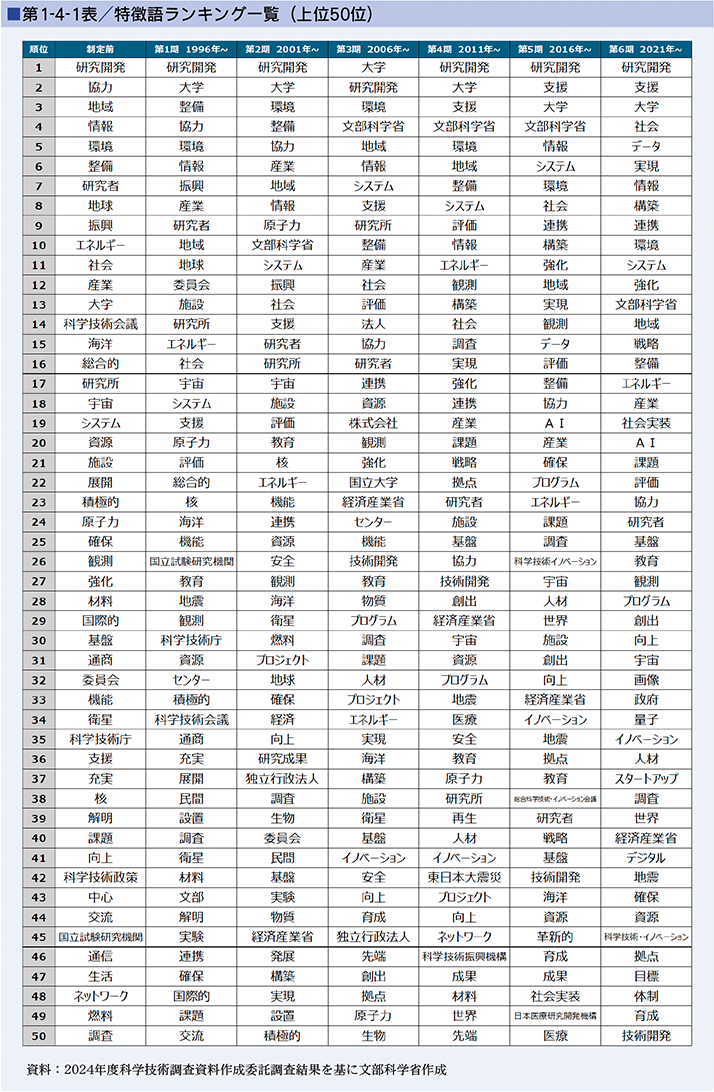}\\[3pt]
  \begin{minipage}{\linewidth}\footnotesize\raggedright
  \textbf{Q:} この表で「AI」は何回記載されていますか？ \emph{(How many times does ``AI'' appear in this table?)}\\
  \textbf{A:} 2回 \emph{(Twice)}
  \end{minipage}
\end{minipage}

\vspace{10pt}

\begin{minipage}[t]{0.32\textwidth}
  \centering
  \qtcard{External knowledge}\\[2pt]
  \qtimg{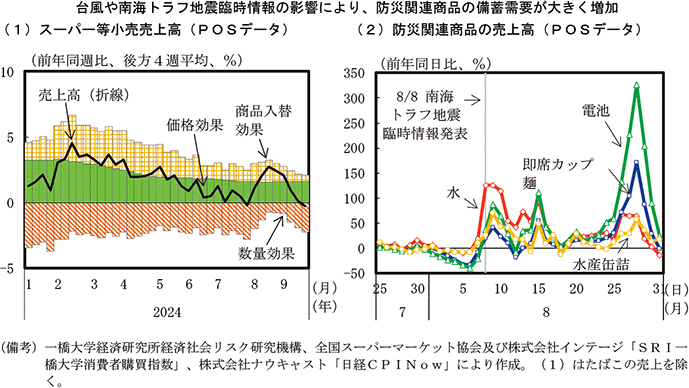}\\[3pt]
  \begin{minipage}{\linewidth}\footnotesize\raggedright
  \textbf{Q:} (2)防災関連商品の売上高を示すグラフにおいて、水及び食料品ではないものは何ですか？ \emph{(In graph (2), which shows sales of disaster-preparedness products, which item is neither water nor a food product?)}\\
  \textbf{A:} 電池 \emph{(Batteries)}
  \end{minipage}
\end{minipage}\hfill
\begin{minipage}[t]{0.32\textwidth}
  \centering
  \qtcard{Visual}\\[2pt]
  \qtimg{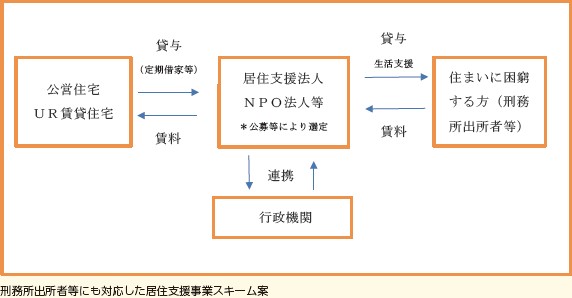}\\[3pt]
  \begin{minipage}{\linewidth}\footnotesize\raggedright
  \textbf{Q:} 右から左に向かう矢印が表しているものは何ですか。 \emph{(What does the arrow pointing from right to left represent?)}\\
  \textbf{A:} 賃料 \emph{(Rent)}
  \end{minipage}
\end{minipage}\hfill
\begin{minipage}[t]{0.32\textwidth}
  \centering
  \qtcard{Other}\\[2pt]
  \qtimg{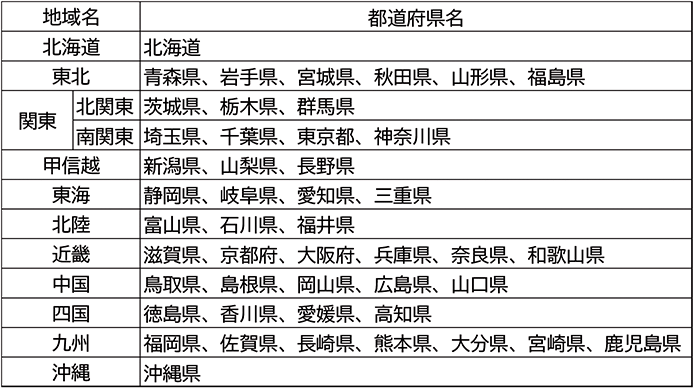}\\[3pt]
  \begin{minipage}{\linewidth}\footnotesize\raggedright
  \textbf{Q:} 示された表において、東海地域の最後に記載されている県はどこですか。 \emph{(In the table shown, which prefecture is listed last within the Tokai region?)}\\
  \textbf{A:} 三重県 \emph{(Mie Prefecture)}
  \end{minipage}
\end{minipage}
\caption{One representative question per question-type flag in \method{}, together with its source image. English glosses are ours; models are evaluated on the original Japanese questions.}
\label{fig:qtype_examples}
\end{figure*}

\section{Accuracy by Image Type and Question Type on \method{}}
\label{sec:appendix_question_type}
\begin{figure}[t]
  \centering
  \includegraphics[width=\linewidth]{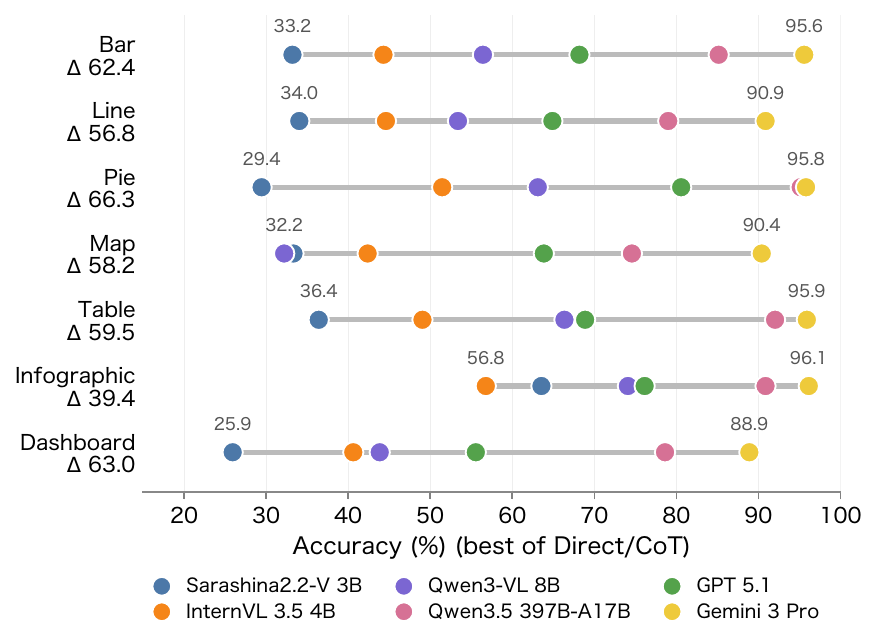}
  \caption{Accuracy spread across models on \method{}, grouped by image type. Categories with fewer than 50 examples (Area, Scatter, Bubble, and Other) are omitted.}
  \label{fig:per_image_type_spread}
\end{figure}

\begin{figure}[t]
  \centering
  \includegraphics[width=\linewidth]{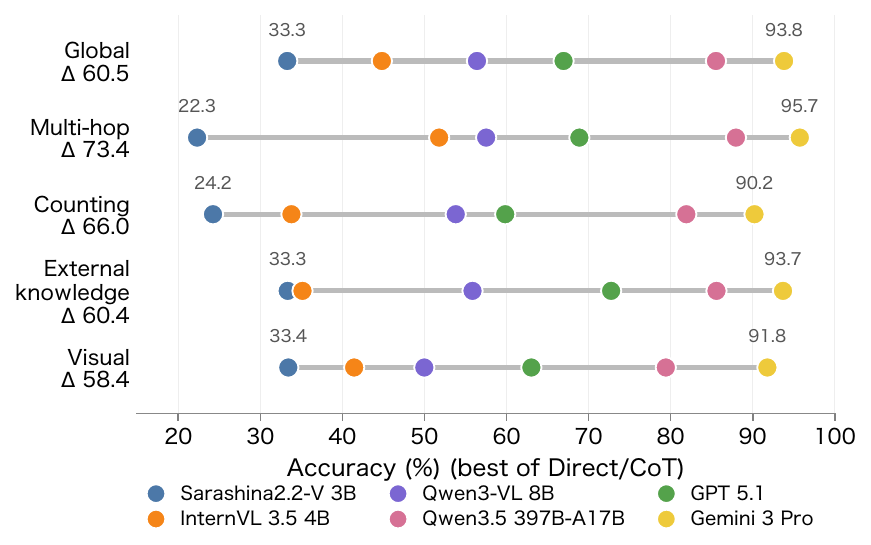}
  \caption{Accuracy spread across models on \method{}, grouped by question type.}
  \label{fig:per_question_type_spread}
\end{figure}

Figures~\ref{fig:per_image_type_spread} and~\ref{fig:per_question_type_spread} show the accuracy spread across models on \method{}, grouped by image type and by question type, respectively. Tables~\ref{tab:per_image_type_acc} and~\ref{tab:per_question_type_acc} report the underlying per-image-type and per-question-type accuracies for a representative subset of models.

\begin{table*}[t]
\centering
\small
\setlength{\tabcolsep}{5pt}
\begin{tabular}{lrrrrrrr}
\toprule
\textbf{Image type} & \textbf{$N$} & \textbf{Sarashina2.2-V} & \textbf{InternVL 3.5} & \textbf{Qwen3-VL} & \textbf{Qwen3.5} & \textbf{GPT 5.1} & \textbf{Gemini 3 Pro} \\
 & & \textbf{3B} & \textbf{4B} & \textbf{8B} & \textbf{397B-A17B} & &  \\
\midrule
Bar & 550 & 33.2 & 44.3 & 56.4 & 85.2 & 68.2 & \textbf{95.6} \\
Line & 281 & 34.0 & 44.6 & 53.4 & 79.0 & 64.9 & \textbf{90.9} \\
Pie & 103 & 29.4 & 51.5 & 63.1 & 95.1 & 80.6 & \textbf{95.8} \\
Map & 59 & 33.3 & 42.4 & 32.2 & 74.6 & 63.8 & \textbf{90.4} \\
Table & 422 & 36.4 & 49.1 & 66.4 & 92.0 & 68.9 & \textbf{95.9} \\
Infographic & 310 & 63.5 & 56.8 & 74.1 & 90.9 & 76.1 & \textbf{96.1} \\
Dashboard & 279 & 25.9 & 40.6 & 43.8 & 78.6 & 55.6 & \textbf{88.9} \\
\bottomrule
\end{tabular}
\caption{Per-image-type accuracy on \method{} (best of Direct/CoT per model). Rows are the image-type categories with at least 50 examples (Section~\ref{sec:exploring}); $N$ is the number of \method{} examples of that type. Each cell is the mean accuracy (\%) over three independent runs. The highest entry in each row is bolded.}
\label{tab:per_image_type_acc}
\end{table*}

\begin{table*}[t]
\centering
\small
\setlength{\tabcolsep}{5pt}
\begin{tabular}{lrrrrrrr}
\toprule
\textbf{Question type} & \textbf{$N$} & \textbf{Sarashina2.2-V} & \textbf{InternVL 3.5} & \textbf{Qwen3-VL} & \textbf{Qwen3.5} & \textbf{GPT 5.1} & \textbf{Gemini 3 Pro} \\
 & & \textbf{3B} & \textbf{4B} & \textbf{8B} & \textbf{397B-A17B} & &  \\
\midrule
Global & 1743 & 33.3 & 44.8 & 56.4 & 85.5 & 67.0 & \textbf{93.8} \\
Multi-hop & 634 & 22.3 & 51.8 & 57.5 & 88.0 & 68.9 & \textbf{95.7} \\
Counting & 293 & 24.2 & 33.8 & 53.8 & 81.9 & 59.8 & \textbf{90.2} \\
External knowledge & 148 & 33.3 & 35.1 & 55.9 & 85.6 & 72.7 & \textbf{93.7} \\
Visual & 897 & 33.4 & 41.4 & 50.0 & 79.4 & 63.0 & \textbf{91.8} \\
\bottomrule
\end{tabular}
\caption{Per-question-type accuracy on \method{} (best of Direct/CoT per model). Rows are the question-type flags defined in Section~\ref{subsec:qa_annotation} (Other omitted); $N$ is the number of \method{} examples whose annotator marked that flag (a single example may contribute to multiple rows). Each cell is the mean accuracy (\%) over three independent runs; the highest entry per row is bolded.}
\label{tab:per_question_type_acc}
\end{table*}

\section{Evaluation Results on Existing Chart and Table Benchmarks}
\label{sec:other_benchmark_results}
\begin{figure*}[t]
  \centering
  \includegraphics[width=\linewidth]{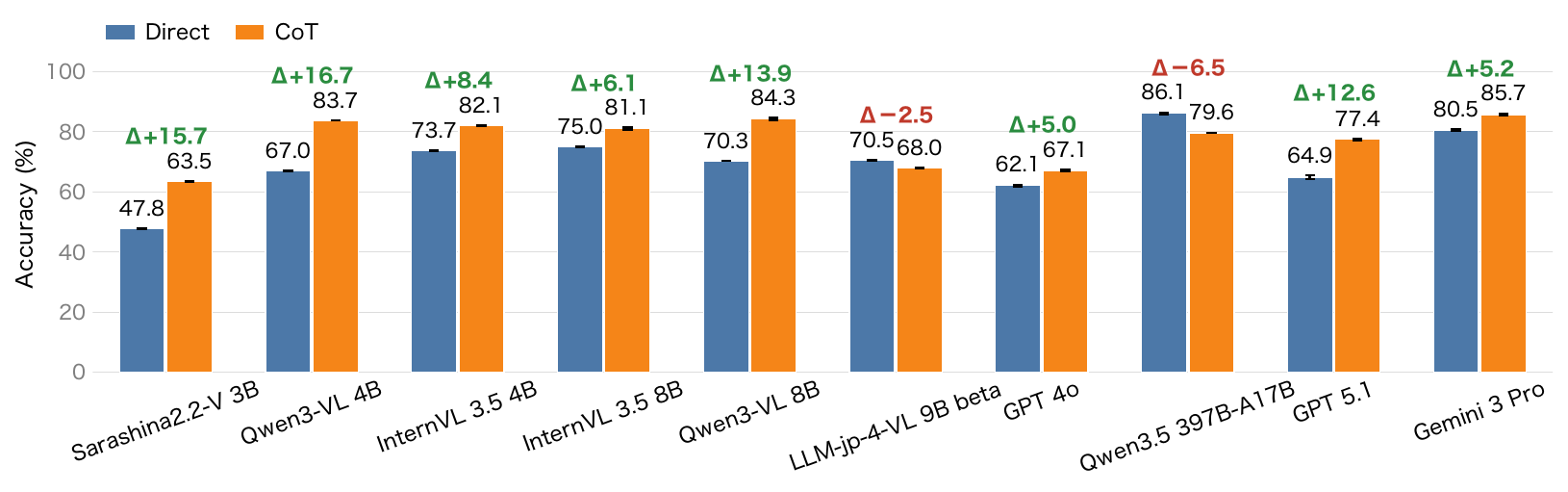}
  \caption{Performance of each model on ChartQA.}
  \label{fig:chartqa_results}
\end{figure*}

\begin{figure*}[t]
  \centering
  \includegraphics[width=\linewidth]{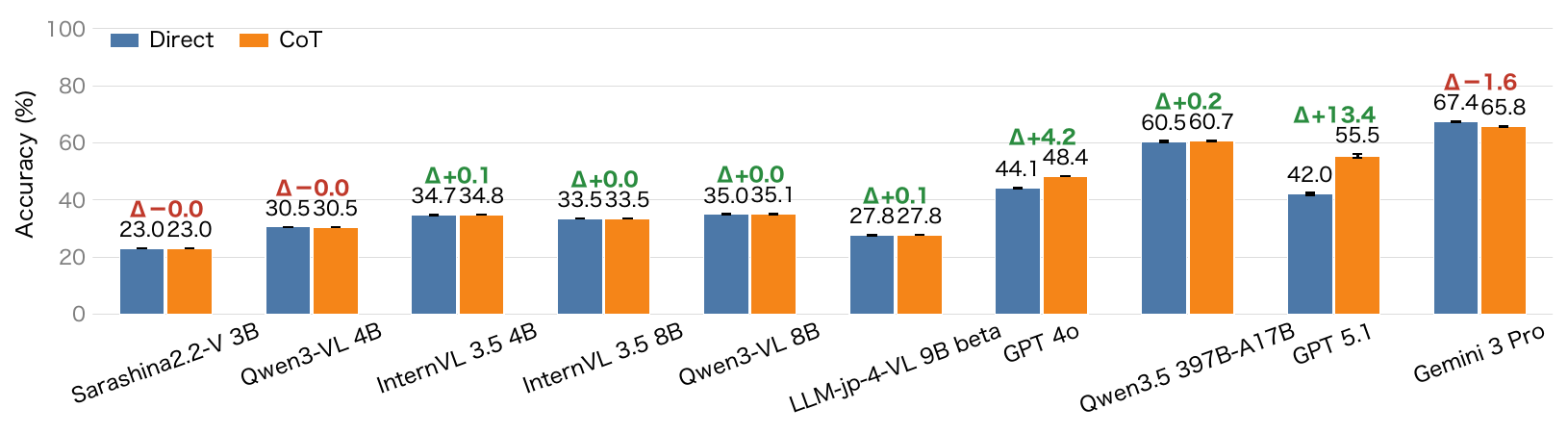}
  \caption{Performance of each model on ChartQAPro.}
  \label{fig:chartqapro_results}
\end{figure*}

\begin{figure*}[t]
  \centering
  \includegraphics[width=\linewidth]{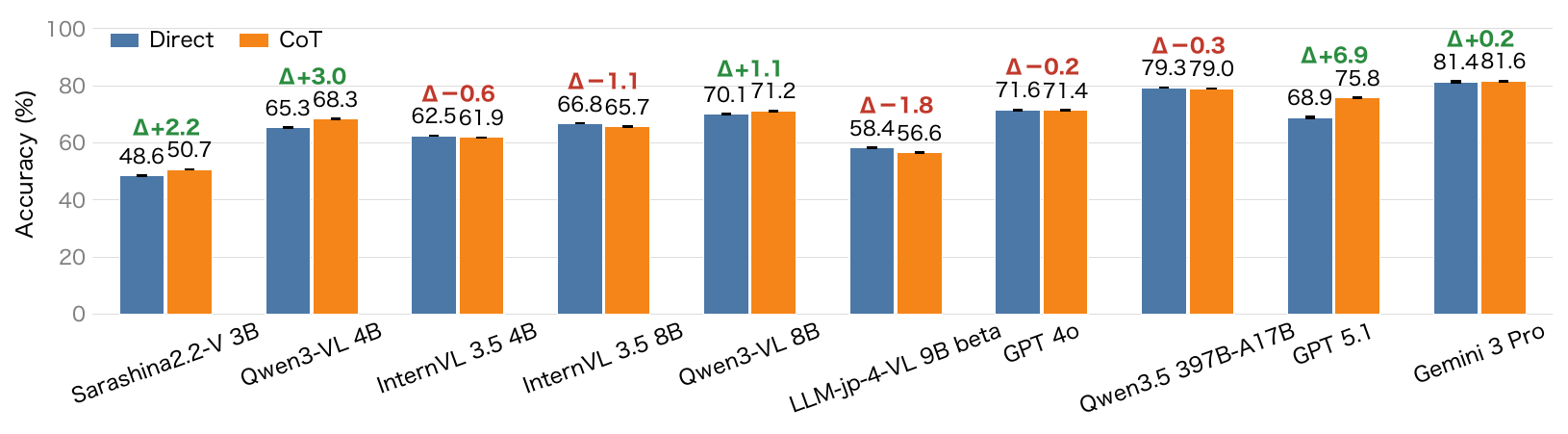}
  \caption{Performance of each model on CharXiv.}
  \label{fig:charxiv_results}
\end{figure*}

\begin{figure*}[t]
  \centering
  \includegraphics[width=\linewidth]{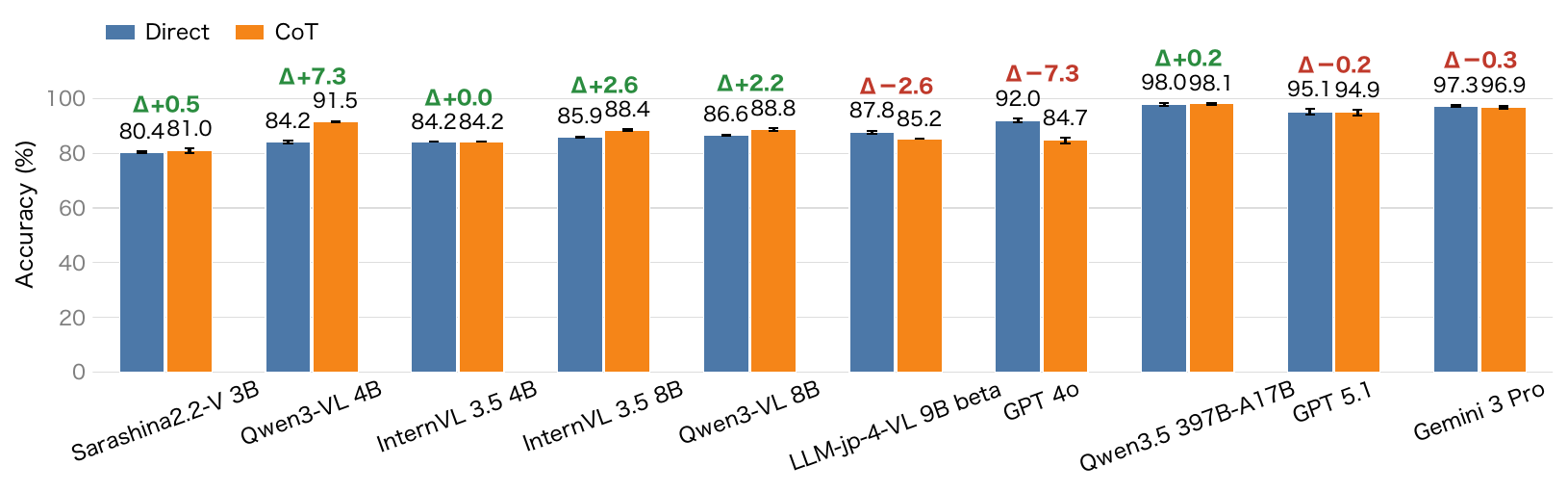}
  \caption{Performance of each model on JGraphQA.}
  \label{fig:jgraphqa_results}
\end{figure*}

Figures~\ref{fig:chartqa_results}, \ref{fig:chartqapro_results}, \ref{fig:charxiv_results}, and~\ref{fig:jgraphqa_results} visualize the per-model Direct and CoT accuracy on ChartQA, ChartQAPro, CharXiv, and JGraphQA as bar charts.

\section{Image Showcases for Comparison Benchmarks}
\label{sec:appendix_showcase_other}

\begin{figure*}[t]
  \centering
  \includegraphics[width=0.5\linewidth]{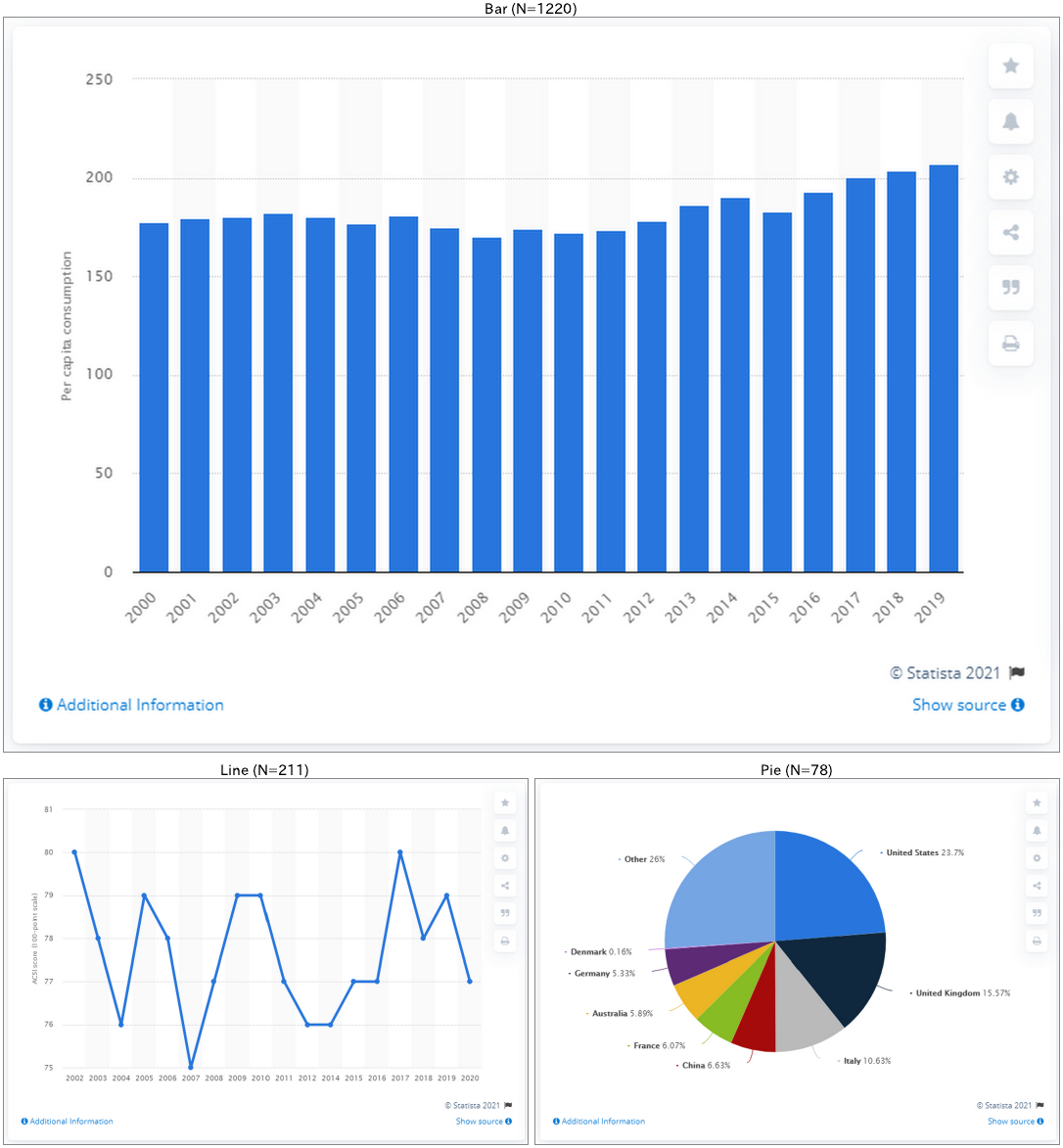}
  \caption{ChartQA: one randomly sampled image per image type.}
  \label{fig:showcase_chartqa}
\end{figure*}

\begin{figure*}[t]
  \centering
  \includegraphics[width=0.5\linewidth]{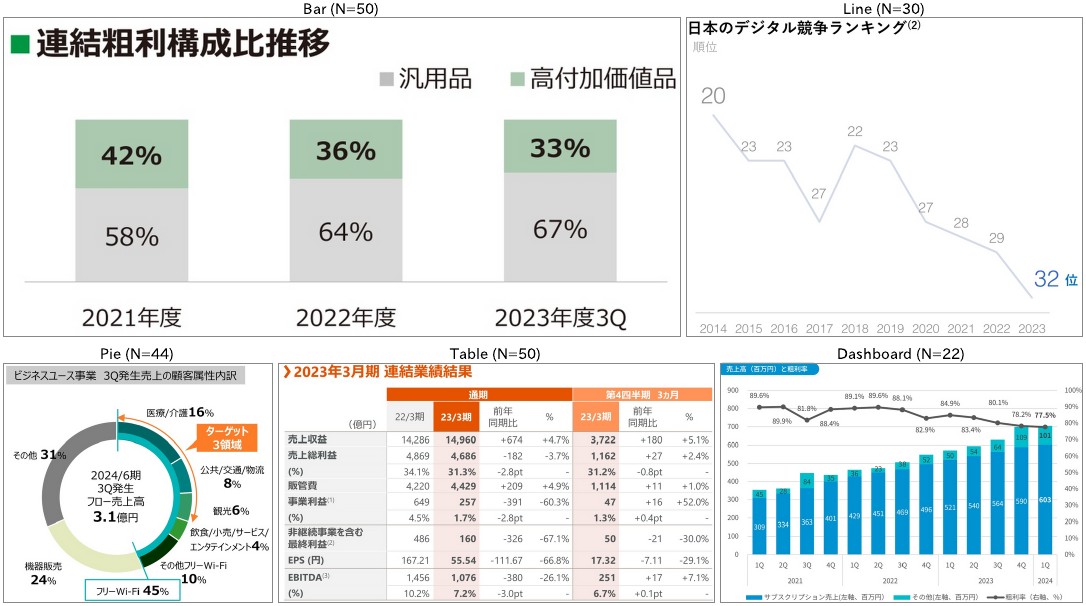}
  \caption{JGraphQA: one randomly sampled image per image type.}
  \label{fig:showcase_jgraphqa}
\end{figure*}

\begin{figure*}[t]
  \centering
  \includegraphics[width=\linewidth]{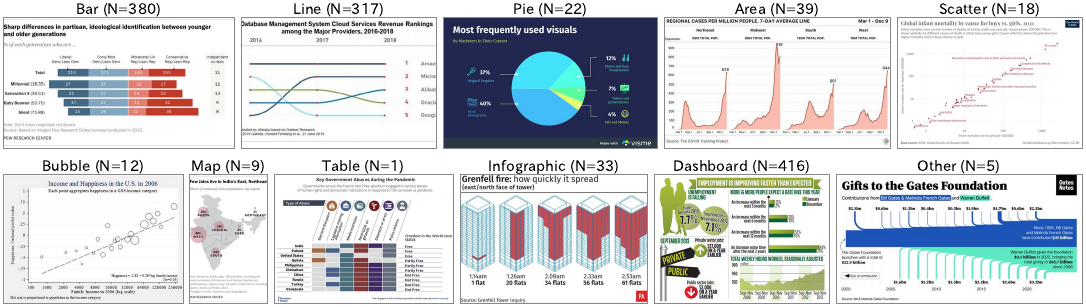}
  \caption{ChartQAPro: one randomly sampled image per image type.}
  \label{fig:showcase_chartqapro}
\end{figure*}

\begin{figure*}[t]
  \centering
  \includegraphics[width=\linewidth]{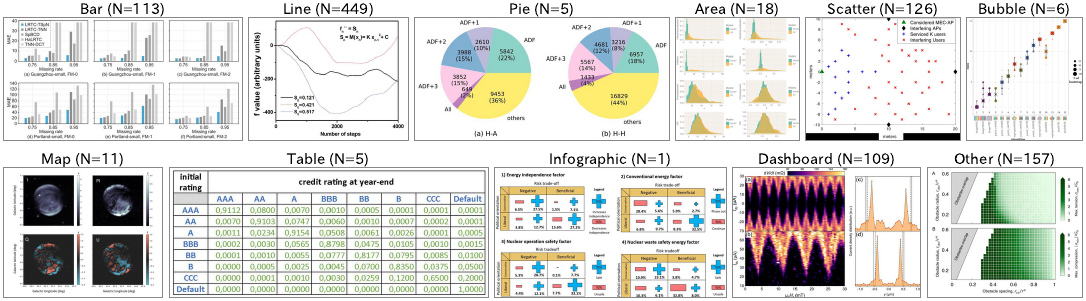}
  \caption{CharXiv: one randomly sampled image per image type.}
  \label{fig:showcase_charxiv}
\end{figure*}

Figures~\ref{fig:showcase_chartqa}, \ref{fig:showcase_jgraphqa}, \ref{fig:showcase_chartqapro}, and~\ref{fig:showcase_charxiv} show one randomly sampled image per image-type category for ChartQA, JGraphQA, ChartQAPro, and CharXiv, respectively.

\end{document}